\documentclass[10pt,journal,cspaper,compsoc]{IEEEtran}

\ifCLASSOPTIONcompsoc
  \usepackage[nocompress]{cite}
\else
  \usepackage{cite}
\fi

\ifCLASSINFOpdf
   \usepackage[pdftex]{graphicx}
   \graphicspath{{../pdf/}{../jpeg/}}
   \DeclareGraphicsExtensions{.pdf,.jpeg,.png,.jpg}
\else
   \usepackage[dvips]{graphicx}
   \graphicspath{{../eps/}}
   \DeclareGraphicsExtensions{.eps}
\fi

\usepackage{amsmath}
\usepackage{amsfonts}

\ifCLASSOPTIONcompsoc
  \usepackage[caption=false,font=footnotesize,labelfont=sf,textfont=sf]{subfig}
\else
  \usepackage[caption=false,font=footnotesize]{subfig}
\fi

\usepackage{stfloats}

\def\ie{{\em i.e.}}
\def\eg{{\em e.g.}}
\def\etal{{\em et al.}}

\newcommand{\figref}[1]{Fig. \ref{#1}}
\newcommand{\tabref}[1]{Tab. \ref{#1}}

\newcommand{\secref}[1]{Section \ref{#1}}

\newcommand{\mc}[1]{\mathcal{#1}}

\newcommand{\br}[1]{\bm{\mathrm{#1}}}

\newcommand{\bs}[1]{\boldsymbol{\texttt{#1}}}

\usepackage{url}
\usepackage{ragged2e}
\usepackage{booktabs}
\usepackage{color}
\usepackage{multirow}
\usepackage{bm}
\usepackage{bbm}
\usepackage[misc]{ifsym}
\usepackage{hyperref}
\usepackage{graphicx}
\usepackage{amsmath}
\usepackage{amsfonts}
\usepackage{enumerate}

\usepackage{balance}

\usepackage{amssymb}

\usepackage[noend]{algpseudocode}
\usepackage{algorithmicx,algorithm}

\graphicspath{{./figures/}}

\begin{document}

\title{
Dual Adaptive Representation Alignment for Cross-domain Few-shot Learning
}

\author{Yifan~Zhao,~\IEEEmembership{Member,~IEEE},~Tong~Zhang,~Jia~Li,~\IEEEmembership{Senior Member,~IEEE},
and~Yonghong~Tian,~\IEEEmembership{Fellow,~IEEE}
\IEEEcompsocitemizethanks{
\IEEEcompsocthanksitem Y. Zhao and T. Zhang contribute equally to this work.
\IEEEcompsocthanksitem Y. Zhao and Y. Tian are with the School of Computer Science, Peking University, Beijing, 100871, China.
\IEEEcompsocthanksitem T. Zhang and J. Li are with the State Key Laboratory of Virtual Reality Technology and Systems, School of Computer Science and Engineering, Beihang University, Beijing, 100191, China.
\IEEEcompsocthanksitem J. Li and Y. Tian are also with the Pengcheng Laboratory, Shenzhen, 518055, China.
\IEEEcompsocthanksitem Y. Tian is also with the School of Electronic and Computer Engineering,
Peking University Shenzhen Granduate School, Shenzhen, 518055, China.
\IEEEcompsocthanksitem J. Li and Y. Tian are the corresponding authors (E-mail: jiali@buaa.edu.cn, yhtian@pku.edu.cn).
}}

\markboth{IEEE TRANSACTIONS ON PATTERN ANALYSIS AND MACHINE INTELLIGENCE}%
{Shell \MakeLowercase{\textit{et al.}}:}

\IEEEtitleabstractindextext{%
\begin{abstract}
\justifying Few-shot learning aims to recognize novel queries with limited support samples by learning from base knowledge. Recent progress in this setting assumes that the base knowledge and novel query samples are distributed in the same domains, which are usually infeasible for realistic applications. Toward this issue, we propose to address the cross-domain few-shot learning problem where only extremely few samples are available in target domains. Under this realistic setting, we focus on the fast adaptation capability of meta-learners by proposing an effective dual adaptive representation alignment approach. In our approach, a prototypical feature alignment is first proposed to recalibrate support instances as prototypes and reproject these prototypes with a differentiable closed-form solution. Therefore feature spaces of learned knowledge can be adaptively transformed to query spaces by the cross-instance and cross-prototype relations. Besides the feature alignment, we further present a normalized distribution alignment module, which exploits prior statistics of query samples for solving the covariant shifts among the support and query samples. With these two modules, a progressive meta-learning framework is constructed to perform the fast adaptation with extremely few-shot samples while maintaining its generalization capabilities. Experimental evidence demonstrates our approach achieves new state-of-the-art results on 4 CDFSL benchmarks and 4 fine-grained cross-domain benchmarks.
\end{abstract}

}

\maketitle

\IEEEdisplaynontitleabstractindextext

\IEEEpeerreviewmaketitle

\IEEEraisesectionheading{\section{Introduction}\label{sec:introduction}}

\IEEEPARstart{S}{upervised} learning with large amounts of annotated data has undoubtedly achieved success in object recognition~\cite{he2016deep,simonyan2014very}, semantic segmentation~\cite{long2015fully} and other computer vision topics~\cite{redmon2016you}. However, such sufficient high-quality annotations are not always available in real-world applications, leading to a huge demand for recognizing objects with few-shot learning~\cite{miller2000learning,fei2006one,lake2015human}. Given sufficient annotated base data, few-shot recognition aims to distinguish objects with novel concepts with limited samples, which presents great challenges to existing supervised learning systems.

To solve this challenge, recent works~\cite{ravi2017optimization,finn2017model,yu2020transmatch} try to extend the base knowledge to novel concepts in a meta-learning manner. These works focus on learning a robust \textit{meta-learner} to extract task-agnostic knowledge and then facilitate the learning of target tasks. For example, Finn~\etal~\cite{finn2017model} propose a Model-Agnostic Meta-Learning (MAML) strategy to learn the initial parameters of the generalized model. Different from these approaches, some ideas~\cite{vinyals2016matching,snell2017prototypical,oreshkin2018tadam,liu2020negative,dhillon2019baseline,tian2020rethinking} tend to solve this problem in a metric-learning manner, which focuses on embedding these samples into a measurable space~\cite{ye2020few,zhang2020deep}. Representative works propose to construct a Matching Net~\cite{vinyals2016matching} or cluster the few-shot samples as Prototypes~\cite{snell2017prototypical,oreshkin2018tadam} to alleviate the outlier instances during distance measurements. Although impressive progress has been achieved, the few-shot learning setting is based on such two hypotheses: 1) the novel dataset for recognition and the base dataset for pretraining are sampled from the same domain; 2) there exists sufficient training samples of base classes in the same domain, which are usually infeasible in realistic industrial applications.

\begin{figure}[!t]
	\centering
	\includegraphics[width=\columnwidth]{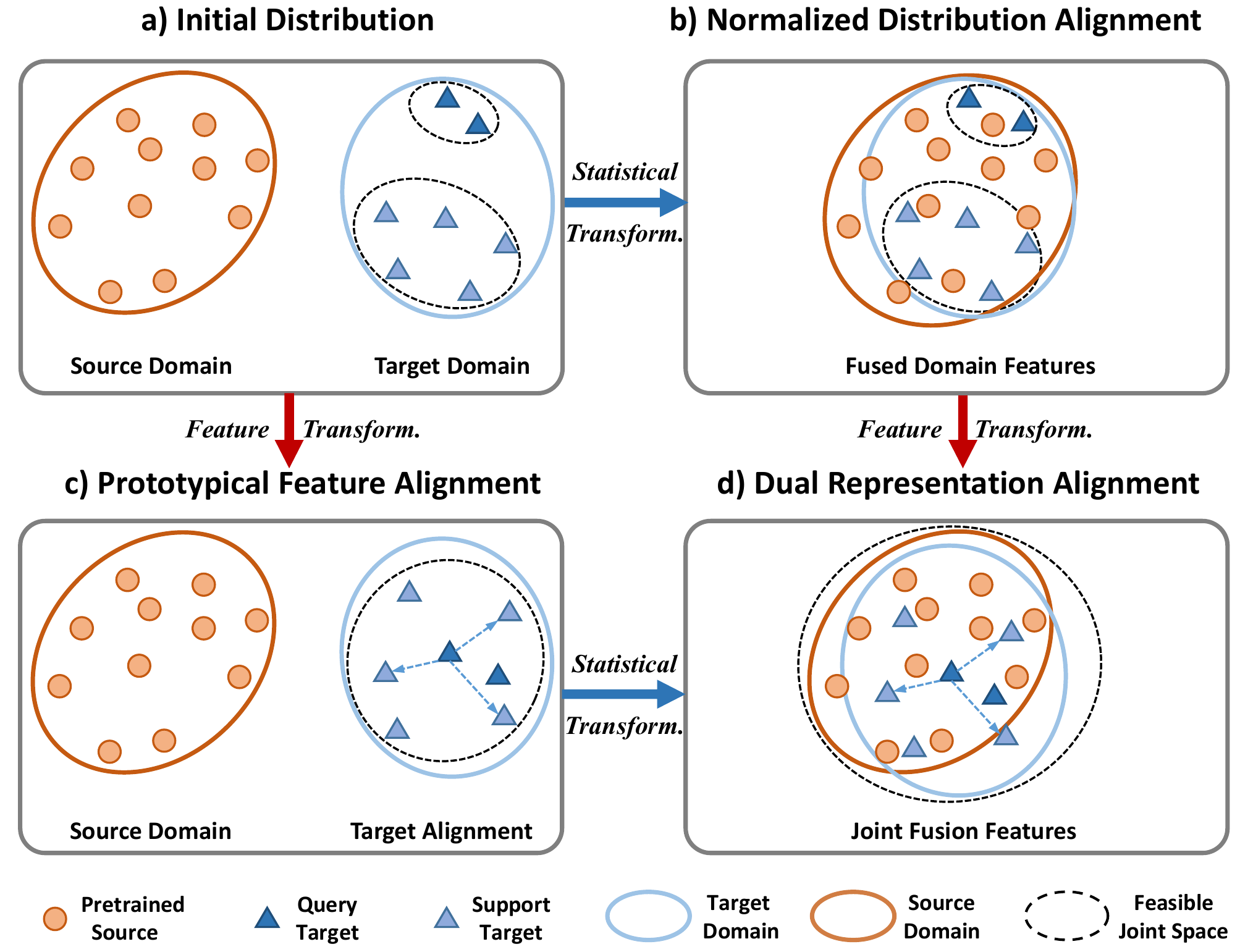}
	\caption{The motivation of proposed approach. Our proposed approach jointly considers the distribution gap with b) normalized distribution alignment and the feature space shifting with c) prototypical feature alignment, forming d) dual representation alignment in one unified learning framework.
	}
	\label{fig:motivation}
\end{figure}

Different from the conventional setting, Cross-Domain Few-Shot Learning (CDFSL)~\cite{guo2020broader} assumes that the pretraining base dataset and the target novel dataset for inference are distributed in different domains, and only few-shot samples in the target domain are available as support set. Prior studies~\cite{chen2019closer,guo2020broader} indicate that existing few-shot learning methods fail to generalize on this cross-domain challenge, which does not take the huge domain gap into consideration. To tackle this new challenging problem, recent efforts learn to construct a generalized feature embedding to ensure the adaptation to target domains, including adopting self-supervised mechanisms~\cite{phoo2021self} and task space augmentations~\cite{liang2021boosting}. Besides, Wang~\etal~\cite{wang2021cross} propose a supervised noise encoder-decoder framework to reconstruct the signal of input images, which aims to form a generalized feature representation with fewer information losses. However, these research efforts only focus on the \textit{generalization capability} of learning models, and the \textit{fast adaptation} with few samples are less considered, which are more crucial issues in cross-domain few-shot learning.

Keeping this in mind, our key idea is to fast adapt a few target queries to existing learning spaces. As in~\figref{fig:motivation} a), the initial distribution shows two major problems: 1) domain shift: the source domain and target domain are sampled from two different distributions; 2) covariate shift: as target samples are extremely rare, target queries for testing usually not fall in the inductive space of target support set. To this end, we propose a Dual adaptive representation alignment (Dara) approach, which is composed of two essential alignment modules,~\ie, normalized distribution alignment and prototypical feature alignment. The first normalized distribution alignment in~\figref{fig:motivation} b) aims to transform the statistics of the source domain aligning with the target domain, hence the query target can be deduced by existing base knowledge. However, although these two distributions are transformed into the same cluster, the huge covariate shift between the query and support targets still exists, which restricts the measurements of novel query classes.

To address this problem, in~\figref{fig:motivation} c), we propose a prototypical feature alignment module to reproject the support prototypes and target queries into another feature space. Beyond ProtoNet~\cite{snell2017prototypical} in few-shot learning, we first present an instance-level recalibration to alleviate the outliers in each prototype. With these recalibrated prototypes, each support prototype is then reprojected into another space based on its relationship to target queries. This new learning space ensures a differentiable closed-form solution with ridge regression~\cite{bertinetto2018meta,wertheimer2021few} to reduce the covariate shifts between query and support features. Similarly, this feature alignment operation also neglects the divergence between query samples and other learnable knowledge,~\ie, pretrained source and target supports.
Hence to jointly consider the feature alignment and distribution alignment in~\figref{fig:motivation} d), our proposed Dara approach not only maintains the generalization capability of meta-learners, but fast aligns the base knowledge to the query target based on its characteristics and relations between them. Moreover, our approach shows robust adaptation ability on different domains with various levels of similarity to natural images.

Contributions of this paper are summarized as:
\begin{enumerate}
	\item[1)]  We propose a novel Dual adaptive representation alignment (Dara) approach for cross-domain few-shot learning, which exploits the fast adaptive alignment based on the characteristics and relations of target queries and base training knowledge.
	\item[2)] We propose a normalized distribution alignment module to align statistics of the source domain with the target domain, and a prototypical feature alignment module to reproject the support and query samples into the same learning space.
	\item[3)] We present a progressive meta-learning framework for cross-domain few-shot tasks, and conduct extensive experiments to demonstrate the robust adaptation capability of our model on diverse levels of benchmarks.
\end{enumerate}

The remainder of this paper is organized as follows:~\secref{sect:relatedwork} reviews related works and~\secref{sect:method} describes the proposed dual adaptive representation alignment approach. Qualitative and quantitative experiments are reported in~\secref{sect:experiment}.~\secref{sect:conclusion} finally concludes the paper.

\section{Related Work} \label{sect:relatedwork}

\subsection{Few-shot Learning}
Recent studies solve this important problem from two different aspects: meta-learning ~\cite{li2017meta, rezende2016one,santoro2016meta,ravi2017optimization,finn2017model,nichol2018first,laenen2021episodes} and metric learning manners~\cite{triantafillou2017few,ren2018meta,oreshkin2018tadam,snell2017prototypical,sung2018learning,vinyals2016matching}. Meta-learning based models aims to learn a \textit{meta-learner} for task agnostic knowledge and then facilitate the learning of target tasks. Some pioneer works~\cite{rezende2016one,santoro2016meta,ravi2017optimization} adopt the recurrent scheme for encoding the few labeled images of novel classes. For example, Rezende~\etal~\cite{rezende2016one} develop a generative algorithm for sequential generation and inference based on the principles of feedback and attention mechanisms, which demonstrate the effectiveness and feasibility of one-shot generation. Tang~\etal~\cite{tang2021mutual} introduce the conditional random field to graph neural models, which helps to implement dependencies in the meta-learning procedure. Other  works~\cite{finn2017model,nichol2018first} propose the model-agonistic meta-learning algorithms which explicitly train the learning model with a limited number of gradient steps, showing the ability for fast adaptation on new novel tasks.

Another line of works~\cite{triantafillou2017few,ren2018meta,oreshkin2018tadam,snell2017prototypical,sung2018learning,vinyals2016matching} tends to solve the few-shot learning task with metric-learning techniques, which constructs an optimal embedding space for distance measurements~\cite{zhou2021binocular,gao2021curvature,ye2020few,zhang2020deep,bertinetto2018meta,wertheimer2021few}. For example, MatchNet~\cite{vinyals2016matching} encodes the few-shot samples with an LSTM network that maps labeled support set with an unlabelled example. ProtoNet~\cite{snell2017prototypical} clusters the support samples and forms an average feature vector as prototypes to alleviate the outliers. Further advanced improvements~\cite{sung2018learning} propose to model a relationship between the query and support samples in the end-to-end network optimization, while TADAM~\cite{oreshkin2018tadam} constructs a task-dependent metric for adaptive measurement based on clustered prototypes. Zhang~\etal~\cite{zhang2020deep} adopt the earth mover's distance as the metric for computing the structure distance between dense image-level representations. Ravi~\etal~\cite{ravi2017optimization} propose an LSTM-based meta-learner to learn the exact optimization algorithm for providing a general initialization for the main learning network. In this paper, we follow the metric learning-based methods to generate adaptive prototypes, which can alleviate the overfitting and quickly adapt to generalize on novel classes.

\begin{figure*}
\begin{center}
\includegraphics[width=1\textwidth]{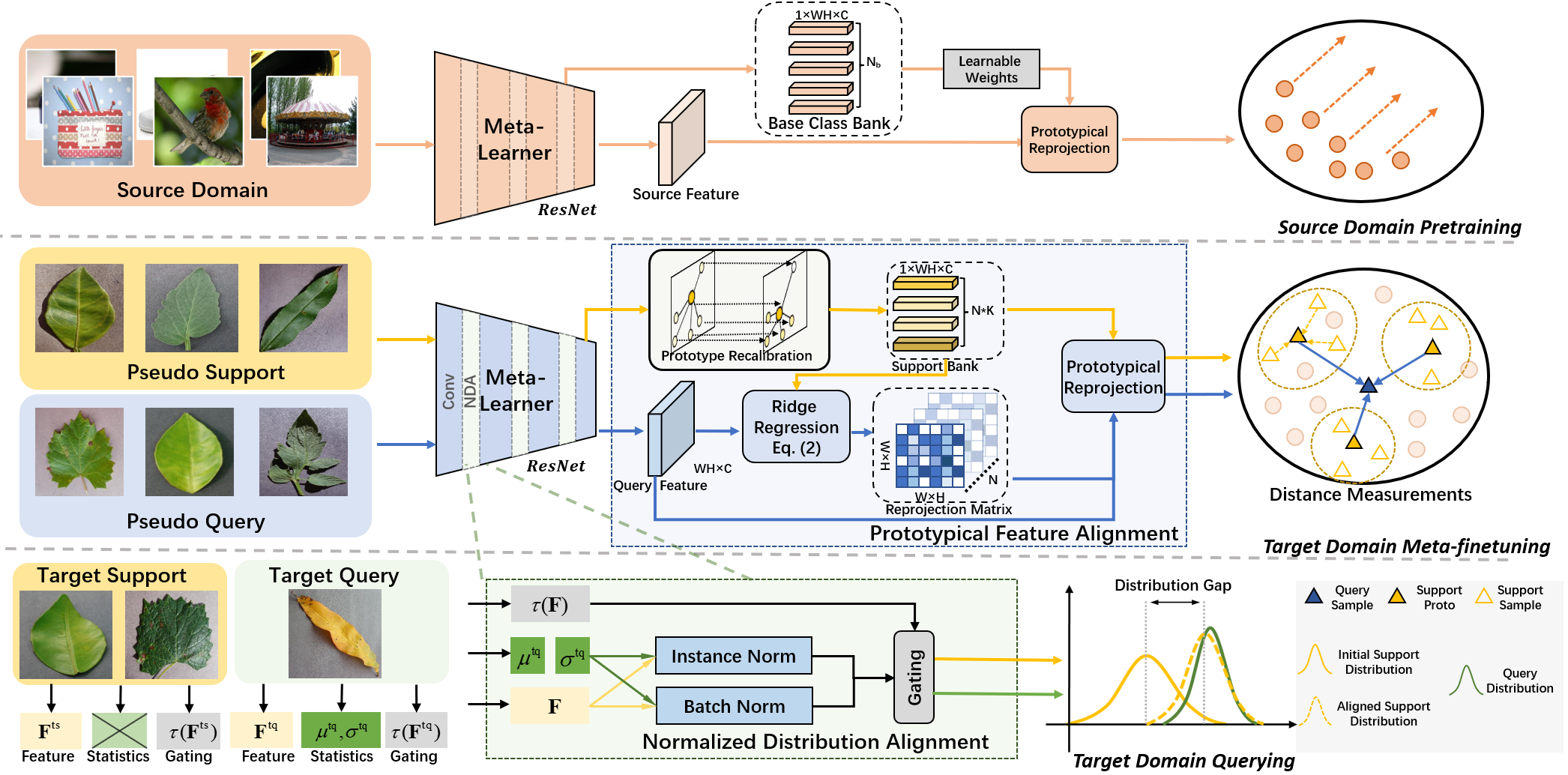}
 \caption{The proposed dual adaptive representation alignment approach consists of two essential modules: 1) prototypical feature alignment exploits the relations across different instances and prototypes for reprojecting learned features and target queries into the same space; 2) normalized distribution alignment focuses on transforming the statistical priors of query knowledge to alleviate the distribution gap between support and query samples. Source domain: miniImageNet~\cite{vinyals2016matching}. Target domain: CropDiseases~\cite{mohanty2016using} for illustration.
 }\label{fig:pipeline}
 \end{center}

\end{figure*}

\subsection{Domain Adaptation}
Deep neural networks trained on the data of source domains face challenges when applied to the target domains. Methods to solve this problem can be divided into two groups. The first group of methods~\cite{chen2018domain,tzeng2014deep,li2017revisiting} tends to solve this problem based on the discrepancy of different domains, while transferring the statistics or network architectures to the target domain. Tzeng~\etal~\cite{tzeng2014deep} propose a Maximum Mean Discrepancy~\cite{borgwardt2006integrating} (MMD) loss constraints to maximize the domain confusions while minimizing the classification error at the same time. Li~\etal~\cite{li2017revisiting} propose to recalculate the batchnorm~\cite{ioffe2015batch} layer to shift the feature-wise distribution to the target domains. The second group of methods~\cite{lee2018diverse,hoffman2018cycada,liu2019few} learns to simulate the target domain with training samples of source domains. Hoffman~\etal~\cite{hoffman2018cycada} propose to solve this problem with adversarial training to maintain the cycle-consistency of different domains. Besides, Liu~\etal~\cite{liu2019few} propose to conduct the image-to-image translation with only few-shot learning samples in an adversarial manner.

\subsection{Domain Problems in Few-shot Learning}

Few research efforts~\cite{guo2020broader,phoo2021self,liu2020feature,tseng2020cross,li2022ranking,das2021confess,memrein,islam2021dynamic} have paid their attention to the domain shifting problems in few-shot learning. Tseng~\etal~\cite{tseng2020cross} construct feature-wise transformations to learn a generalized distribution that can adapt to any unseen domains. Cross-Domain Few-Shot Learning~\cite{guo2020broader} is introduced as a challenge and establishes the BSCD-FSL benchmark with a collection of various datasets, where the base and novel classes are drawn from different domains and span over plant diseases, satellite, and medical images.
With this new challenging problem, several works~\cite{chen2019closer,guo2020broader,phoo2021self,liang2021boosting} show that existing few-shot methods fail to generalize on this cross-domain task. Phoo~\etal~\cite{phoo2021self} propose a self-training procedure with unlabeled data from target domains, which constructs a teacher model with distillation techniques. Islam~\etal~\cite{islam2021dynamic} propose a dynamic distillation-based approach that uses the unlabeled data to learn cross-domain feature representations. While Xu~\etal~\cite{memrein} also focus on the distillation method with a memorized and restitution strategy for discriminative information learning. However, these existing techniques~\cite{guo2020broader,phoo2021self} do not regularize the task learning with well-distributed hyper-planes, and neglect the domain shifts of source domains~\cite{deng2009imagenet} for pretraining and target domains~\cite{mohanty2016using,helber2019eurosat,tschandl2018ham10000,codella2019skin,wang2017chestx} for inference, leading to an incomplete usage of existing network knowledge.

\subsection{Discussions and Relations}
As aforementioned, this paper focuses on the cross-domain few-shot learning problem, which assumes that: 1) the source knowledge and inference query samples are from two distinct distributions, and 2) there only exists limited \textit{samples} in the target domains. Existing few-shot learning works,~\eg, ProtoNet~\cite{snell2017prototypical}, MatchNet~\cite{vinyals2016matching}, are theoretically based on the hypothesis that the base samples and query samples are sampled from the same distribution, which neglects to discover the domain shifts in real applications. Besides these few-shot learning methods, this cross-domain problem also stands at a different perspective from other domain adaptation settings~\cite{chen2018domain,tanwisuth2021prototype}, including the few-shot domain adaptation~\cite{yue2021prototypical}, which learns with limited labels but sufficient samples in target domains.
To solve this less-explored task, we propose a novel dual adaptive representation alignment approach, which focuses on the adaptation ability by utilizing two fast transformation cues: the feature-wise relationship between the query and support samples and the distribution gap among query samples and existing knowledge.

\section{Method}\label{sect:method}
In this section, we introduce a dual adaptive representation alignment approach for cross-domain few-shot learning, which is composed of two essential modules in three learning stages. As in~\figref{fig:pipeline}, we first present a prototypical feature alignment module in~\secref{sec:protoalign} to recalibrate the sample features within each prototype and to reproject the prototypes by their relationship to query features. Besides this improvement, we then introduce a normalized distribution alignment module in~\secref{sec:normalign} to shift the base knowledge to target queries. With these two effective modules, in~\secref{sec:learning}, we present a progressive meta-learning framework to optimize this learning process for fast adaptation.
\subsection{Problem Formulation}
Cross-domain few-shot learning problem is derived from the basic few-shot learning setting, which aims to recognize novel concepts based on the existing base knowledge. Let $\mc{X}$ and $\mc{Y}$ denote the image space and label space and $(\br{x},\br{y}) \in \mc{X} \times \mc{Y} $ be the input and its corresponding label. The major difference between these two tasks is that the cross-domain setting assumes that the target novel samples and base samples are distributed in two different domains,~\ie, $(\br{x}^s,\br{y}^s) \in \mc{S}, (\br{x}^t,\br{y}^t) \in \mc{T}$ and $\mc{P}(\mc{S}) \ne \mc{P}(\mc{T})$ with distribution function $\mc{P}$. Based on the learning of source domain $\mc{S}$, the target domain $\mc{T}= \{\mc{T}^s,\mc{T}^q\}$ is further divided into two sub-tasks for support $\mc{T}^s$ and query distribution $\mc{T}^q$, where sample pairs are notated as $\{\br{x}^{ts},\br{y}^{ts}\}_{i=1}^{K\times N} \in \mc{T}^s$ and $\{\br{x}^{tq},\br{y}^{tq}\}_{i=1}^{M} \in \mc{T}^q$ respectively. The label space $\mc{Y}^{tq}$ is disjoint from $\mc{Y}^{ts}$.  Following the original work~\cite{guo2020broader}, the learning process follows a meta-learning trend with an $N$-way $K$-shot labeled subset for support, which is available for target domain tuning, and then constructs a generalized representation to measure the distance of $M$ query samples.

\begin{figure}[!t]
	\centering
	\includegraphics[width=\columnwidth]{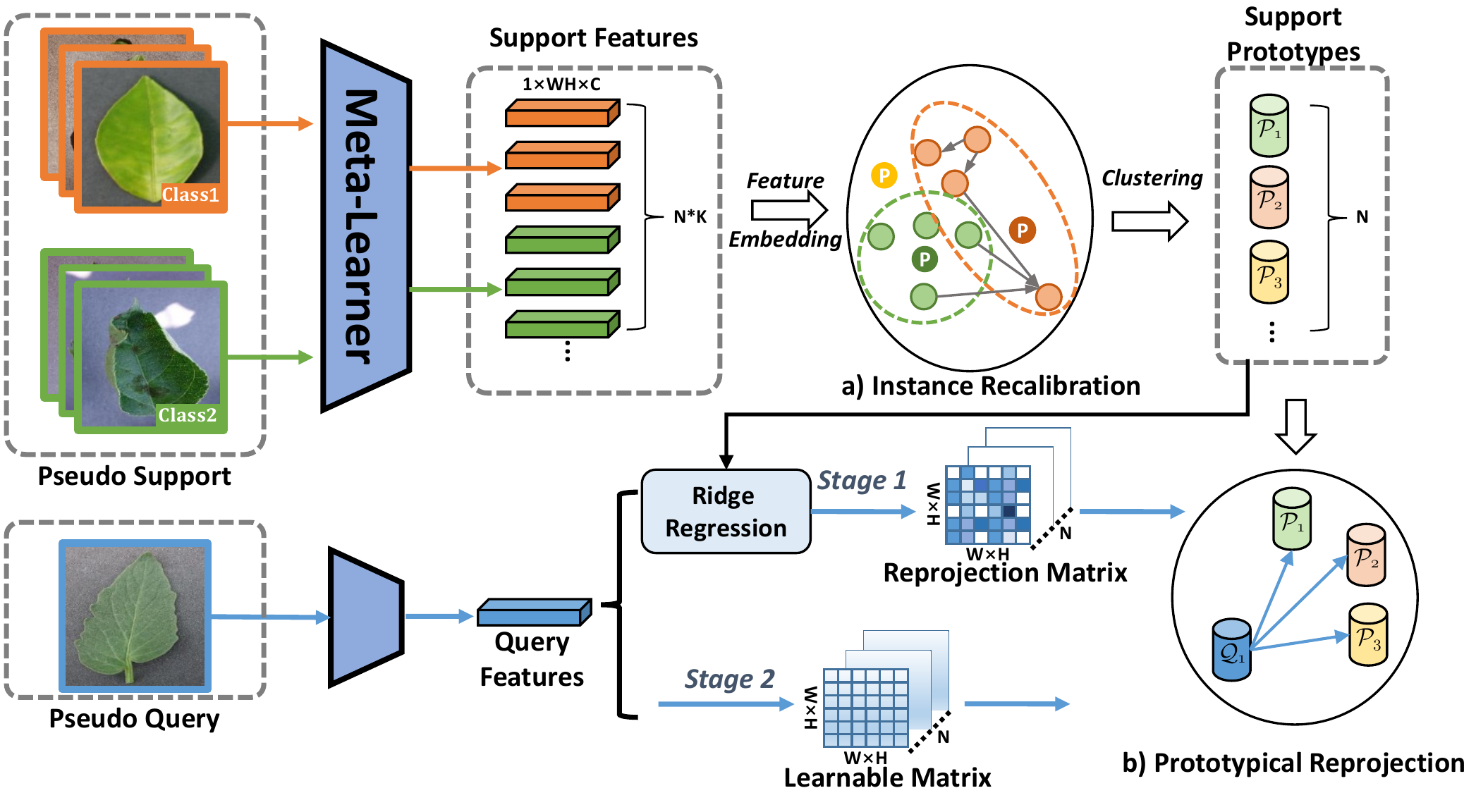}
	\caption{Illustration of Prototypical Feature Alignment (PFA) module. Our PFA module first exploits the instance-wise relationships to recalibrate different support samples and conducts a closed-form regression by the relations among query and support prototypes. }	\label{fig:proto}
\end{figure}

\subsection{Prototypical Feature Alignment} \label{sec:protoalign}

Given the input samples $\br{x}$ with a meta-learning feature extractor $\br{F}=\Phi(\br{x}) \in \mathbb{R}^{W\times H\times C}$, the most common way for measuring the distance of query and support samples is to form prototype embedding~\cite{snell2017prototypical}. Prototype networks first aggregate the support samples of the $n$th category as a unified cluster $\br{P}_n=\frac{1}{K}\sum_{k=1}^{K} \br{F}_{n,k}$ and then measure the distance between query samples and category prototypes.

\textbf{Instance-level feature recalibration.} Meta-learning methods follow an episodic training procedure, in which samples support and query instances from a large-scale dataset, which can be deemed as a uniformed distribution space and shows strong generalization capability. However, in this cross-domain setting, the learnable samples in the target domains $\mc{T}^s$ are extremely limited,~\eg, only 25 samples for 5-way 5-shot learning. This would lead to an obvious issue: during the finetuning on target domains, learning systems always construct fixed prototypes using all these limited supported samples, and outlier samples would cause a severe inductive bias in the gradient descent process.

To this end, in~\figref{fig:proto}, we first establish a meta-finetuning strategy and randomly split target support samples $\mc{T}^s$ into two parts, $\mc{T}^s=\{ \widetilde{\mc{T}}^{ps},\widetilde{\mc{T}}^{pq} \}$ for pseudo support and pseudo query in each episode. Each set follows the standard way-shot task split,~\ie, $N$-way $K_{ps}$-shot for $\widetilde{\mc{T}}^{ps}$.
In this manner, we then introduce an adaptive instance-level feature recalibration operation to reweight each sample based on the overall similarities to other samples:
\begin{equation}\label{eq:recali}
\begin{split}
\br{A}_{n,i} &= \frac{1}{|\widetilde{\mc{T}}^{ps}|-1}\sum_{\forall \br{x}_{n,j} \in \widetilde{\mc{T}}^{ps}} ^ {\br{x}_{n,i} \ne \br{x}_{n,j}} \frac{\br{F}_{n,i}^ \top  \cdot \br{F}_{n,j}}{{\left\| {\br{F}_{n,i}} \right\| \left\| {\br{F}_{n,j}} \right\|}}, \br{F}=\Phi(\br{x}),\\
\br{P}_n & =\frac{1}{|\widetilde{\mc{T}}^{ps}|} \sum_{k=1}^{|\widetilde{\mc{T}}^{ps}|} \br{A}_{k} \cdot \br{F}_{n,k}, n=1\ldots N,
\end{split}
\end{equation}
where each feature $\br{F}_{n,i}$ is vectorized with the size of $\mathbb{R}^{1 \times WHC}$.
In Eqn.~\eqref{eq:recali}, $\br{A} \in \mathbb{R}^{N\times K_{ps}}$ is the adaptive weight calculated by the similarity scores of one instance to others in the pseudo support set. Despite its simplicities, this operation can easily construct a pair-wise relationship and recalibrate the instances for correct prototypes.

\textbf{Prototypical feature reprojection.} After obtaining the established prototypes, the standard way is to measure the distance between query samples and support prototypes. As the samples are very limited in the target domain, there usually exists a covariant shift between support and query samples. Hence our goal is to reproject the prototypes to another measuring space where the query samples are also distributed. Following this idea, we introduce the differentiable regression manner to build reprojection matrices $\br{W} \in \mathbb{R}^{R\times KR}, R=W\times H$ for learning this embedding.
Pioneer works~\cite{bertinetto2018meta,wertheimer2021few} have shown their effectiveness by using a closed-form solution for prototypical reprojection, which performs the conventional Ridge regression:
\begin{equation}\label{eq:ridge}
\br{W}^*= \mathop{\arg\min}_{\br{W}} ||\br{W}\br{P}_{i}- \br{Q} ||^2 + \lambda ||\br{W}||^2,
\end{equation}
where $\br{Q}=\Phi(\br{x}^{pq}) \in \mathbb{R}^{R\times C}$ denotes the extracted features of pseudo query samples $\br{x}^{pq}$ by the same meta-learner. Following~\cite{wertheimer2021few}, we keep the spatial dimension $R$ of feature maps for learning a concrete reprojection. With closed-form solution for Eqn.~\eqref{eq:ridge}, prototypes $\hat{\br{P}}$ are reprojected as:
\begin{equation}\label{eq:reproj}
\hat{\br{P}}= \br{W}^*\br{P}=\br{Q}\br{P}^\top (\br{P}\br{P}^\top +\lambda \br{I})^{-1} \br{P} \in \mathbb{R}^{KR\times C}.
\end{equation}
This means the projection for each prototype is constrained by two factors: the distribution space of query feature $\br{Q}$ and the regularization term controlled by $\lambda$.

Given random pseudo query samples $\br{x}^{pq} \in \widetilde{\mc{T}}^{pq}$ and pseudo support samples $\br{x}^{ps} \in \widetilde{\mc{T}}^{ps}$ in each episode, the overall measurements $\mc{M}(\cdot)$ with a learnable temperature $\gamma$, following~\cite{wertheimer2021few,dhillon2019baseline}, can be presented as:
\begin{equation}\label{eq:measure}
\mc{M}(\br{W}^*\br{P}_i,\br{Q}) = \frac{e^{-\frac{\gamma}{R}||\br{W}^*\br{P}_i-\br{Q}||^2 }} {\sum_{j=1}^{N} e^{-\frac{\gamma}{R}||\br{W}^*\br{P}_j-\br{Q}||^2 }}.
\end{equation}
Hence the distance of query feature $\br{Q}$ and reprojected prototypes $\br{W}^*\br{P}$ can be formulated as a probability distribution. With this closed-form solution, our learning objective first trains the parameters $\theta$ in meta-learners $\Phi$:
\begin{equation}\label{eq:meta}
\mathop{\arg\min}_{\theta} \mathbb{E}_{(\widetilde{\mc{T}}^{ps},\widetilde{\mc{T}}^{pq}) \sim \mc{T}^s} \mc{L}^{m}(\mc{M}(\br{W}^*\Phi_{\theta}(\br{x}^{ps}),\Phi_{\theta}(\br{x}^{pq})); \br{y}^{pq}),
\end{equation}
where $\mc{L}^{m}$ denotes the meta-learning loss (\textit {Stage 1} in~\figref{fig:proto}), which is a standard cross-entropy loss of $N$ classes.

During this pseudo episodic training process, the meta-learner converges with optimized parameters $\hat{\theta}$ to adapt to target domains.
However, the reprojection matrices $\br{W}^*$ are still fixed in this meta-finetuning, which is only the quasi-optimal solution for embedding owing to the matrix inversion operation. We therefore use learnable matrices $\mc{R}(\br{Z})=\br{Q}\br{Z}^\top (\br{Z}\br{Z}^\top +\lambda \br{I})^{-1} \br{Z}$ instead of the calculated $\br{W}^*\br{P}$, and only finetune the reprojection prototypes $\br{Z}$ with the fixed feature extractor $\Phi_{\hat{\theta}}$:
\begin{equation}\label{eq:catmat}
\mathop{\arg\min}_{\br{Z}} \mathbb{E}_{(\widetilde{\mc{T}}^{ps},\widetilde{\mc{T}}^{pq}) \sim \mc{T}^s} \mc{L}^{z}(\mc{M}(\mc{R}(\br{Z}),\Phi_{\hat{\theta}}(\br{x}^{pq})); \br{y}^{pq}),
\end{equation}
where $\mc{L}^{z}$ is the cross-entropy loss which only optimizes matrices $\br{Z}$ (\textit {Stage 2} in~\figref{fig:proto}). Different from previous methods~\cite{bertinetto2018meta,wertheimer2021few}, the reason for our stage-wise training mechanism is that when jointly optimizing $\br{Z}$ and meta-learner $\Phi_{\theta}$, reprojection prototypes $\br{Z} \in \mathbb{R}^{R\times C}$ are only deemed as a special classification head with MLPs for $N$ categories. Hence we first obtain a quasi-optimal reprojection $\br{W}^*$ for learning generalized representations of meta-learners and then finetune the reprojection $\mc{R}(\br{Z})$ to adapt to target domains.

\begin{figure}[!t]
	\centering
	\includegraphics[width=\columnwidth]{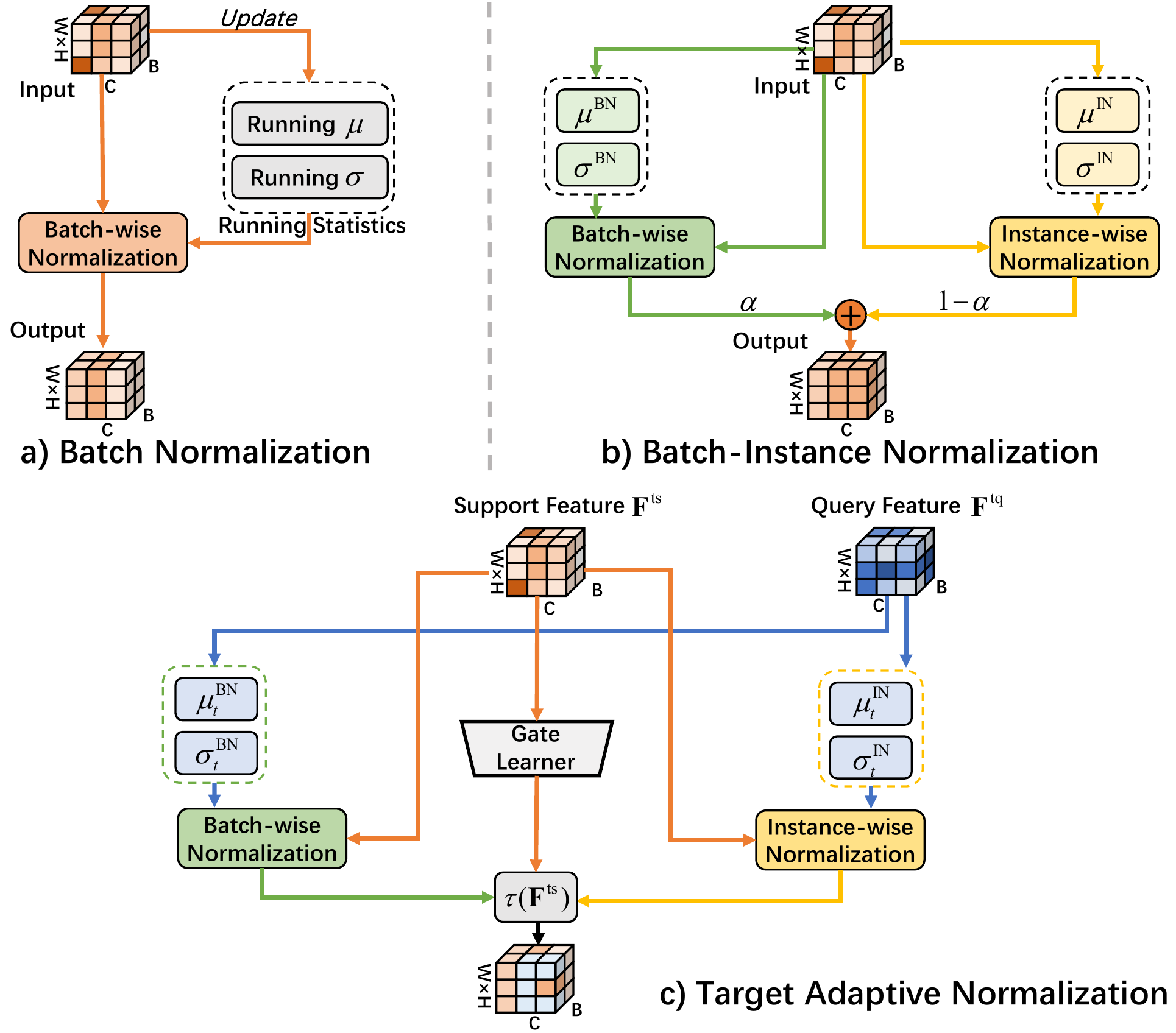}
	\caption{Illustration of different normalization strategies. a) Conventional batch normalization~\cite{ioffe2015batch} with running statistics. b) BIN~\cite{nam2018batch} with fused batch~\cite{ioffe2015batch} and instance normalization~\cite{ulyanov2016instance}. c) Ours: transforming the statistics of query features for support feature space, while learning an feature-related gating strategy.  }	\label{fig:norm}
\end{figure}

\subsection{Normalized Distribution Alignment}\label{sec:normalign}
Besides the feature alignment, the other problem for fast domain adaptation is that the query samples are distributed in different subspaces from both target support and source samples. As the available instances in target domains are extremely limited, conducting distribution alignments using learning methods would also lead to severe overfitting issues. Hence we propose a target adaptive normalization by directly learning statistic priors from query samples.

\textbf{Preliminaries.} Conventional Batch Normalization (BN)~\cite{ioffe2015batch} or Instance Normalization (IN)~\cite{ulyanov2016instance} follows a minibatch learning trend, which constrains feature distributions to a standard space. Let $\br{F} \in \mathbb{R}^{B\times W'H'\times C'}$ denote the input feature of certain layer for easy notation, batch normalization~\cite{ioffe2015batch} operation in~\figref{fig:norm} a) has the following form:
\begin{equation}\label{eq:bn}
\begin{split}
\br{F}_{BN}& = \frac{\br{F}-\mu_{BN}} { \sqrt{\sigma_{BN}^{2} +\epsilon}}, \\
\mu_{BN} &= \frac{1}{BW'H'}\sum_{b=1}^{B}\sum_{h=1}^{H'}\sum_{w=1}^{W'} \br{F}_{b,c,w,h} \in \mathbb{R}^{ C'},\\
\sigma^{2}_{BN} &= \frac{1}{BW'H'}\sum_{b=1}^{B}\sum_{h=1}^{H'}\sum_{w=1}^{W'} (\br{F}_{b,c,w,h}-\mu_{BN})^2.
\end{split}
\end{equation}
Similarly, Instance Normalization (IN)~\cite{ulyanov2016instance} is proposed to learn the per-instance statistics, whose mean $\mu_{IN}$ and $\sigma_{IN}$ are calculated over the resolution dimensions:
\begin{equation}\label{eq:in}
\begin{split}
\mu_{IN} &= \frac{1}{W'H'}\sum_{h=1}^{H'}\sum_{w=1}^{W'} \br{F}_{b,c,w,h} \in \mathbb{R}^{B\times C'},\\
\sigma^{2}_{IN} &= \frac{1}{W'H'}\sum_{h=1}^{H'}\sum_{w=1}^{W'} (\br{F}_{b,c,w,h}-\mu_{IN})^2.
\end{split}
\end{equation}
In this manner, the instance-wise style differences are also removed in IN operation, resulting in a better measurable feature $\br{F}_{IN}$ for classification. Hence BIN~\cite{nam2018batch} in~\figref{fig:norm} b) is proposed to balance the style-invariant features (IN) and the style-related features (BN):
\begin{equation}\label{eq:bin}
\br{F}_{BIN} = \alpha \cdot \br{F}_{IN} + (1-\alpha)\cdot \br{F}_{BN},
\end{equation}
where $\alpha \in [0,1]$ to balance this fusion operation and the following affine weights are omitted for easy notation.

\textbf{Target adaptive normalization.} Although BIN~\cite{nam2018batch} has enhanced style-invariant features for classification embedding, the query samples are usually distributed in different subspaces which diverge hugely from the support ones. This covariant shift is very hard to obviate by existing domain generalization techniques. To this end, our goal is to transform the distribution of target support $\br{F}^{ts}$ to align with the query ones $\br{F}^{tq}$, based on statistics of query features:
\begin{equation}\label{eq:tan}
\begin{split}
\br{H}_{TAN}^{ts} & =\mc{N}^{T}_{BN}(\br{F}^{ts};\br{F}^{tq})+\mc{N}^{T}_{IN}(\br{F}^{ts};\br{F}^{tq}) \\
& = \frac{\br{F}^{ts}-\mu_{BN}(\br{F}^{tq})} { \sqrt{\sigma_{BN}^{2}(\br{F}^{tq}) +\epsilon}} +
\frac{\br{F}^{ts}-\mu_{IN}(\br{F}^{tq})} { \sqrt{\sigma_{IN}^{2}(\br{F}^{tq}) +\epsilon}},
\end{split}
\end{equation}
where $\sigma(\cdot),\mu(\cdot)$ denote calculating the mean and variance statistics of input features. Moreover, as IN and BN play different roles across different datasets and even instances, hence in~\figref{fig:norm} c), we propose to learn the adaptive function $\tau(\cdot)$ to balance the fusion weight:
\begin{equation}\label{eq:balance}
\begin{split}
\widetilde{\br{H}}_{TAN}^{ts}=(1-\tau(\br{w};\br{F}^{ts}) )\cdot \mc{N}^{T}_{BN}(\br{F}^{ts};\br{F}^{tq})\\
+\tau(\br{w};\br{F}^{ts})\cdot \mc{N}^{T}_{IN}(\br{F}^{ts};\br{F}^{tq}).
\end{split}
\end{equation}
This formula constraints to learn an adaptive fusion strategy $\tau(\cdot)$, which should be learned using target support samples $\br{F}^{ts}$. The typical manner to implement $\tau(\cdot)$ is using single parameter $\br{w}$ or linear transformation~\ie, $\tau(\br{F})=\bs{Sigmoid} (\br{w}\cdot \bs{Pool}(\br{F})+\br{b})$ with learnable weights $\br{w}$ only related to $\br{F}^{ts}$. During the inference time for different queries, the learnable $\tau(\cdot)$ is fixed. The final feature $\widetilde{\br{H}}_{TAN}^{ts}$ shows two distinct advantages under this extreme few-shot setting: 1) transforming the support samples to query distributions with the statistic prior guidance; 2) learning an instance-aware fusion to enhance the adaptive representation of different normalization strategies.

\subsection{Progressive Meta-Learning Framework}\label{sec:learning}
As illustrated in~\figref{fig:pipeline}, in our framework, we summarize the cross-domain few-shot learning process into three progressive procedures.
\begin{enumerate}
\item \textbf{Source domain pretraining:} given the large pretraining source dataset,~\ie, miniImageNet~\cite{vinyals2016matching}, and let $(\br{x}^s,\br{y}^s) \in \mc{S}$ be any sampled paired data, our aim in this phase is to learn a generalized feature extractor $\Phi_\theta(\cdot)$. We first randomly initialized base prototypes $\br{P}$ for all $N$ base classes, and then conduct the standard learning process following the measurements in previous studies~\cite{wertheimer2021few} to obtain the generalized network parameter $\theta^{p}$.
\item \textbf{Target domain meta-finetuning:} after the pretraining stage, the meta-learner $\Phi_\theta(\cdot)$ is initialized with pretraining parameters $\theta^{p}$. With this feature extractor, we split the target support samples into the pseudo query and pseudo support to conduct meta-learning on target domains. In this manner, we first perform an instance-level feature recalibration based on the relationships of support instances. After that, our optimization includes two stages: i) in the first stage, we optimize meta-learner parameters $\theta$ with a calculated reprojection weight $\br{W}^{*}$, which reflects the relations of query samples and support samples; ii) in the second stage, the meta-learner are fixed with learned $\hat \theta$ and then we perform optimization on reprojection weights $\br{Z}$.
\item \textbf{Target domain querying:} With the optimized $\hat{\br{Z}}$ and $\hat \theta$, in the querying phase, we repeat the above feature extraction operations and prototypical feature construction operation, and then calculate the statistics of each query instance,~\ie, $\mu(\br{F}^{tq}),\sigma(\br{F}^{tq})$. Then we inject the query statistics to embed the support samples for distribution normalization. The final measurement scores $\br{S}$ can be obtained using transformed support and original queries.
\end{enumerate}

\begin{algorithm}[!t]
\caption{Progressive Meta-Learning}\label{alg:pml}
\hspace*{0.02in} {\bf Input:}
Source domain data $(\br{x}^{s},\br{y}^{s}) \in \mc{S}$, Base class num $N^s$ , target domain support set $\{\br{x}^{ts},\br{y}^{ts}\}_{i=1}^{K\times N} \in \mc{T}^s$, target domain query set $\{\br{x}^{tq},\br{y}^{tq}\}_{i=1}^{M} \in \mc{T}^q$.  \\
\hspace*{0.02in} {\bf Output:}
Meta-learner: $\theta$ for feature extractor $\Phi(\cdot)$, $\br{Z}$ for reprojection in Eqn. ({\color{red}3}), measured Scores $\br{S}$.
\begin{algorithmic}[1]
\State Init. meta-learner: $\theta$ in $\Phi(\cdot)$ with random norm

\Comment{{\color{blue}\textit{Source Domain Pretraining}}}
\State Random Init. base prototypes: ${\br{P}} \in \mathbb{R}^{N^s \times R \times C}$
\For{ $\forall (\br{x}^{s},\br{y}^{s})  \in \mc{S}$}
\State Extracte features from meta-learner:
\Statex \qquad $\br{F}^{s}=\Phi(\br{x}^{s}) \in \mathbb{R}^{W\times H\times C}$
\State Calculate $\mc{M}(\br{F}^{s},\br{P})$ by Eqn. ({\color{red}4})
\State Optimize $\theta^{p},\br{P}^{*}  = \mathop{\arg\min}_{\theta,\br{P}} \mc{L}^{CE}(\mc{M}(\cdot))$
\EndFor

\Comment{{\color{blue}\textit{Target Domain Meta-finetuning}}}

\State Init. meta-learner $\Phi(\cdot)_{\theta}$ using $\theta^{p}$
\For{ $ \forall \{ (\br{x}^{ts},\br{y}^{ts}) \}_{i=1}^{K\times N}  \in \mc{T}^s$}
\State Split $\{ (\br{x}^{ts},\br{y}^{ts}) \}_{i=1}^{K\times N}$ as pseudo support set $\{(\br{x}^{ps},\br{y}^{ps})\}$ and pseudo query set $\{(\br{x}^{pq},\br{y}^{pq})\}$
\State Extracte features $\br{F}^{ps},\br{F}^{pq}$ using $\Phi(\cdot)_{\theta}$
\State Recalibrate support set using Eqn. ({\color{red}1})
\State Calculate $\br{W}^{*}$ using Eqn. ({\color{red}2}) by $\br{F}^{ps},\br{F}^{pq}$
\State Calculate pseudo support protos:
\Statex \qquad $\br{P}^{ps}=\frac{1}{K}\sum_{i=1}^{K} \br{F}^{pq}_{n,i}$
\State Reproject pseudo support protos $\hat{\br{P}}^{ps}$ by Eqn. ({\color{red}3})
\State Calculate $\mc{M}(\br{F}^{pq},\hat{\br{P}}^{ps})$ by Eqn. ({\color{red}4})
\State Optimize $\hat{\theta} = \mathop{\arg\min}_{\theta} \mc{L}^{m}(\mc{M}(\cdot;\br{Z}))$ in Eqn. ({\color{red}5})  with fixed $\mc{R}{(\br{Z})}=\br{W}^*$
\State Optimize $\hat{\br{Z}} = \mathop{\arg\min}_{\br{Z}} \mc{L}^{w}(\mc{M}(\cdot; \hat \theta))$ in Eqn. ({\color{red}6})  with fixed $\hat \theta$
\EndFor

\Comment{{\color{blue}\textit{Target Domain Querying}}}

\For{ $ \{ (\br{x}^{tq},\br{y}^{tq}) \}  \in \mc{T}^q$, $\{\br{x}^{ts},\br{y}^{ts}\}_{i=1}^{K\times N} \in \mc{T}^s$}
\State Repeat Step {\color{red}10} to {\color{red}13} using optimized $\hat \theta, \hat{\br{Z}}$
\State Calculate query statistics $\mu(\br{F}^{tq}),\sigma(\br{F}^{tq})$
\State Normalize $\br{H}_{TAN}^{ts},\br{H}_{TAN}^{tq}$ by Eqn. ({\color{red}10})
\State Measure scores $\br{S}$ of $\br{H}_{TAN}^{ts},\br{H}_{TAN}^{tq}$ using Eqn. ({\color{red}4})
\EndFor
\State \Return Measured Scores $\br{S}$ of $N$ classes
\end{algorithmic}
\end{algorithm}

With this meta-Learning framework, the existing seen knowledge, including statistical prior and network parameters, can be progressively transferred to target queries with few-shot limited samples.

\begin{table*}[t]
\setlength{\tabcolsep}{0.7mm}
\caption{5-way k-shot classification accuracy on miniImageNet $\to$ BSCD-FSL dataset. Mean accuracies and 95\% confidence interval are reported. Top half: training under the original setting~\cite{guo2020broader}. Bottom half: training under using additional augmentations.
 * means the results are reported by the original BSCD-FSL benchmarks~\cite{guo2020broader}. $\br{A}$: learning with augmented larger datasets. $\br{L}$: using label propagation in test queries. Higher performance under the same setting is bolded for a better view.}
\label{tab:comp}
\renewcommand{\arraystretch}{1.2}
\centering
\resizebox{\textwidth}{!}{
\begin{tabular}{c ccc|ccc|ccc|ccc}
\toprule[1.5pt]
\multirow{1}{*}{}& \multicolumn{3}{c}{CropDiseases}& \multicolumn{3}{c}{EuroSAT} & \multicolumn{3}{c}{ISIC} & \multicolumn{3}{c}{ChestX} \\
\cline{2-13}
Methods&1-shot& 5-shot& 20-shot & 1-shot& 5-shot& 20-shot& 1-shot& 5-shot& 20-shot & 1-shot& 5-shot& 20-shot \\
\midrule[1.5pt]
MAML*~\cite{finn2017model}&- &78.05$\pm$0.68 &89.75$\pm$0.42 &- &71.70$\pm$0.72 &81.95$\pm$0.55 &- &40.13$\pm$0.58 &52.36$\pm$0.57 &- &23.48$\pm$0.96 &27.53$\pm$0.43 \\
Proto*~\cite{snell2017prototypical} &- &79.72$\pm$0.67 &88.15$\pm$0.51 &- &73.29$\pm$0.71 &82.27$\pm$0.57 &- &39.57$\pm$0.57 &49.50$\pm$0.55 &- &24.05$\pm$1.01 &28.21$\pm$1.15 \\
Proto+FWT*~\cite{tseng2020cross}&- &72.72$\pm$0.70 &85.82$\pm$0.51 &- &67.34$\pm$0.76 &75.74$\pm$0.70 &- &38.87$\pm$0.52 &43.78$\pm$0.47 &- &23.77$\pm$0.42 &26.87$\pm$0.43 \\
Finetune~\cite{guo2020broader} &- &89.25$\pm$0.51 &95.51$\pm$0.31  &- &79.08$\pm$0.61 &87.64$\pm$0.47  &-  &48.11$\pm$0.64 &59.31$\pm$0.48  &- &25.97$\pm$0.41 &31.32$\pm$0.45\\
TransFT~\cite{guo2020broader}&- &90.64$\pm$0.54 &95.91$\pm$0.72 &- &81.76$\pm$0.48 &87.97$\pm$0.42 &- &49.68$\pm$0.36 &61.09$\pm$0.44 &- &26.09$\pm$0.96 &31.01$\pm$0.59\\
TPN-ATA~\cite{wang2021cross} &\multicolumn{1}{l}{77.82$\pm$0.5}&\multicolumn{1}{l}{88.15$\pm$0.5}\ &- &\multicolumn{1}{l}{65.94$\pm$0.5}& \multicolumn{1}{l}{79.47$\pm$0.3}&-&\multicolumn{1}{l}{34.70$\pm$0.4} &\multicolumn{1}{l}{45.83$\pm$0.3}&- & \multicolumn{1}{l}{21.67$\pm$0.2}&\multicolumn{1}{l}{23.60$\pm$0.2}&-\\
STARTUP~\cite{phoo2021self} &75.93$\pm$0.80&93.02$\pm$0.45&97.51$\pm$0.21 &63.88$\pm$0.84& 82.29$\pm$0.60& 89.26$\pm$0.43&32.66$\pm$0.60 &47.22$\pm$0.61& 58.63$\pm$0.58&\textbf{23.09$\pm$0.43}&26.94$\pm$0.44& 33.19$\pm$0.46\\
NSAE~\cite{liang2021boosting} &-&93.31$\pm$0.42& 98.33$\pm$0.18 &-&84.33$\pm$0.55 & 92.34$\pm$0.35 &-&55.27$\pm$0.62&67.28$\pm$0.61&-&27.30$\pm$0.45&35.70$\pm$0.47\\
Dara (Ours) &\textbf{80.74$\pm$0.76}&\textbf{95.32$\pm$0.34}&\textbf{98.58$\pm$0.15}& \textbf{67.42$\pm$0.80} &\textbf{85.84$\pm$0.54}&\textbf{93.14$\pm$0.30} &\textbf{36.42$\pm$0.64}&\textbf{56.28$\pm$0.66} &\textbf{67.36$\pm$0.57} & 22.92$\pm$0.40& \textbf{27.54$\pm$0.42} &\textbf{35.95$\pm$0.47} \\
\midrule[1.5pt]
LMMPQS ($\br{A}$)~\cite{yeh2020large}  &-&93.52$\pm$0.39& 97.60$\pm$0.23 &-&86.30$\pm$0.53 & 92.59$\pm$0.31 &-&51.88$\pm$0.60&64.88$\pm$0.58&-&26.10$\pm$0.44&32.58$\pm$0.47\\
BSR ($\br{A}$)~\cite{liu2020feature}  &-&93.99$\pm$0.39& 98.62$\pm$0.15 &-&82.75$\pm$0.55 & 92.61$\pm$0.31 &-&54.97$\pm$0.68&66.43$\pm$0.57&-&28.20$\pm$0.46&\textbf{36.72$\pm$0.51}\\
ConFeSS ($\br{A}$)~\cite{das2021confess} &-&\multicolumn{1}{l}{88.88}&\multicolumn{1}{l}{95.34}\  &-&\multicolumn{1}{l}{84.65}& \multicolumn{1}{l}{90.40}&-&\multicolumn{1}{l}{48.85} &\multicolumn{1}{l}{60.10}&- & \multicolumn{1}{l}{27.09}&\multicolumn{1}{l}{33.57}\\
NSAE ($\br{A}$+$\br{L}$)~\cite{liang2021boosting}  &-&96.09$\pm$0.35& 99.20$\pm$0.14 &-&87.53$\pm$0.50 & 94.21$\pm$0.29 &-&56.85$\pm$0.67&67.45$\pm$0.60&-&28.73$\pm$0.45&36.14$\pm$0.50\\
RDC-FT ($\br{A}$+$\br{L}$)~\cite{li2022ranking}  &\multicolumn{1}{l}{\textbf{86.33$\pm$0.5}}&\multicolumn{1}{l}{93.55$\pm$0.3}\ &- &\multicolumn{1}{l}{\textbf{71.57$\pm$0.5}}& \multicolumn{1}{l}{84.67$\pm$0.3}&-&\multicolumn{1}{l}{35.84$\pm$0.4} &\multicolumn{1}{l}{49.06$\pm$0.3}&- & \multicolumn{1}{l}{22.27$\pm$0.2}&\multicolumn{1}{l}{25.48$\pm$0.2}&-\\
Dara (Ours) ($\br{A}$) &81.50$\pm$0.66&\textbf{96.23$\pm$0.34} &\textbf{99.21$\pm$0.11}& 69.39$\pm$0.84&\textbf{87.67$\pm$0.54}& \textbf{94.40$\pm$0.27} &\textbf{38.49$\pm$0.66}&\textbf{57.54$\pm$0.68} &\textbf{68.43$\pm$0.54} &\textbf{22.93$\pm$0.40}& \textbf{28.78$\pm$0.45} &36.20$\pm$0.43 \\
\bottomrule[1.5pt]
\end{tabular}
}
\end{table*}

\begin{table*}[t]
\setlength{\tabcolsep}{0.7mm}
\caption{5-way k-shot classification accuracy on miniImageNet $\to$ fine-grained benchmark datasets. Mean accuracies and 95\% confidence interval are reported. The training settings follow the original setting~\cite{guo2020broader} on these datasets.
 $\dag$ means the results are reported by~\cite{liang2021boosting}.  Higher performance in the same setting is bolded for a better view.}
\label{tab:comp_fine}
\renewcommand{\arraystretch}{1.3}
\centering
\resizebox{\textwidth}{!}{
\begin{tabular}{c ccc|ccc|ccc|ccc}
\toprule[1.5pt]
\multirow{1}{*}{}& \multicolumn{3}{c}{Car~\cite{krause20133d}}& \multicolumn{3}{c}{CUB~\cite{wah2011caltech}} & \multicolumn{3}{c}{Plantae~\cite{van2018inaturalist}} & \multicolumn{3}{c}{Places~\cite{zhou2017places}} \\
\cline{2-13}
Methods&1-shot& 5-shot& 20-shot & 1-shot& 5-shot& 20-shot& 1-shot& 5-shot& 20-shot & 1-shot& 5-shot& 20-shot \\
\midrule[1.5pt]
Finetune~\cite{guo2020broader}\dag &- &52.08$\pm$0.74 &79.27$\pm$0.63  &- &64.14$\pm$0.77 &84.43$\pm$0.65  &-  &59.27$\pm$0.70 &75.35$\pm$0.68  &- &70.06$\pm$0.74 &80.96$\pm$0.65\\
TPN-ATA~\cite{wang2021cross} &\multicolumn{1}{l}{34.18$\pm$0.4}&\multicolumn{1}{l}{46.95$\pm$0.4}\ &- &\multicolumn{1}{l}{50.26$\pm$0.5}& \multicolumn{1}{l}{65.31$\pm$0.4}&-&\multicolumn{1}{l}{39.83$\pm$0.4} &\multicolumn{1}{l}{55.08$\pm$0.3}&- & \multicolumn{1}{l}{\textbf{57.03$\pm$0.5}}&\multicolumn{1}{l}{72.12$\pm$0.4}&-\\
NSAE~\cite{liang2021boosting} &-&54.91$\pm$0.74& 79.68$\pm$0.54 &-&68.51$\pm$0.76 & 85.22$\pm$0.56 &-&59.55$\pm$0.74&75.70$\pm$0.64&-& 71.02$\pm$0.72&82.70$\pm$0.58\\
BSR~\cite{liu2020feature}\dag  &-&57.49$\pm$0.72& \textbf{81.56$\pm$0.78} &-&69.38$\pm$0.76 & 85.84$\pm$0.79 &-&61.07$\pm$0.76&77.20$\pm$0.90&-&71.09$\pm$0.68&81.76$\pm$0.81\\
Dara (Ours) &\textbf{35.25$\pm$0.57}&\textbf{58.44$\pm$2.39} &81.38$\pm$0.59& \textbf{52.70$\pm$0.83}&\textbf{77.51$\pm$0.65}& \textbf{89.60$\pm$0.43} &\textbf{42.08$\pm$0.55}&\textbf{65.40$\pm$1.95} &\textbf{79.54$\pm$0.64} &51.25$\pm$0.58&\textbf{72.15$\pm$0.43} &\textbf{83.12$\pm$0.54} \\
\bottomrule[1.5pt]
\end{tabular}
}
\end{table*}

\section{Experiments}\label{sect:experiment}
\subsection{Experiment Settings}
\textbf{Dataset and benchmarks.} Guo~\etal~\cite{guo2020broader} first set up a Cross-Domain Few-Shot Learning (CDFSL) benchmark. Following this CDFSL benchmark, which spans over different challenges for cross-domain transformation, we adopt 4 standard datasets reflecting the real-world applications for few-shot learning: 1) CropDiseases~\cite{mohanty2016using}: natural images of agricultural diseases; 2) EuroSAT~\cite{helber2019eurosat}:  satellite images which captured in different views but their colors are still natural; 3) ISIC~\cite{tschandl2018ham10000,codella2019skin}: skin diseases with lost perspective distortions; 4) ChestX~\cite{wang2017chestx}: gray images of different splanchnic diseases which are most dissimilar with the natural images. Note that the above dataset also has huge amounts of unlabelled data in the target domain, which we do not take as unsupervised training data.

For the fine-grained setting of cross-domain few-shot learning, following previous work~\cite{wang2021cross,liang2021boosting,tseng2020cross}, we also conduct verifications of model generalization capabilities on the widely-used fine-grained datasets~\ie, CUB~\cite{wah2011caltech}, Stanford-Cars~\cite{krause20133d}, Places~\cite{zhou2017places} and Plantae~\cite{van2018inaturalist}. Different from the datasets adopted in CDFSL~\cite{guo2020broader}, these fine-grained datasets consist of natural fine-grained categories and share similarities with the pretraining dataset miniImageNet~\cite{vinyals2016matching}.

\textbf{Evaluation protocols.} For all datasets, we take miniImageNet~\cite{vinyals2016matching} as natural base knowledge for source training data and then use the above datasets for finetuning. Following prior works~\cite{phoo2021self,wang2021cross}, we consider 1-shot, 5-shot, and 20-shot settings with 5-ways for testing, where larger shot numbers would be better learned by supervised training techniques. We follow the benchmark evaluation protocols~\cite{guo2020broader}, which simulates 600 independent few-shot episodes, and report the average accuracy with 95\% confidence intervals.

\textbf{Implementation details.} To perform a fair comparison with existing works~\cite{guo2020broader,phoo2021self,wang2021cross,liang2021boosting,tseng2020cross},  we perform all experiments using the same ResNet10 backbone~\cite{he2016deep} with SGD optimizers in PyTorch framework. We perform a pretraining on miniImageNet~\cite{vinyals2016matching} with a learning rate of 0.05 and batch size of 64 for 350 epochs, and conduct meta-finetuning on each target domain for 100 epochs (50 epochs for each stage in~\secref{sec:protoalign}) with a learning rate of 0.01. The input image resolution is $160\times 160$, leading to a small resolution in the final feature map,~\ie, $W=H=5$.  In each meta-finetuning episode, the numbers of pseudo query $\widetilde{\mc{T}}^{pq}$ and support shots $\widetilde{\mc{T}}^{ps}$ are set as $1$ and $|\mc{T}^{s}|-1$ respectively. For the 1-shot learning setting, we take this labeled sample as both the pseudo query and pseudo support set, which can be deemed as a self-reconstruction process. Code is available at \url{https://github.com/iCVTEAM/Dara}.

\subsection{Comparison with State of the Arts}

\textbf{Vanilla setting for cross-domain few-shot learning.} In the benchmark setting without additional data augmentations, existing methods on this CDFSL task can be divided into two categories,~\ie, transfer learning methods~\cite{guo2020broader,liu2020feature} and meta-learning methods~\cite{finn2017model,snell2017prototypical,wang2021cross,liang2021boosting}. As reported in~\tabref{tab:comp}, state-of-the-art methods for general few-shot learning, including ProtoNet~\cite{snell2017prototypical}, MAML~\cite{finn2017model}, have achieved inferior results to learn a generalized embedding when compared to the simple finetuning strategy~\cite{guo2020broader},~\eg, 79\% accuracy for ProtoNet while 89\% for finetuning on the simple CropDiseases~\cite{mohanty2016using} dataset. This indicates that these state-of-the-art few-shot learning methods perform bad feature extraction on novel tasks when facing severe domain-shifting issues.

Recent ideas~\cite{liang2021boosting,phoo2021self,wang2021cross} propose to learn robust feature generalization for domain transferring. For example, STARTUP~\cite{phoo2021self} shows a notable higher performance than existing learning methods~\cite{guo2020broader},~\ie, 93.02\% on CropDiseases, while it takes large amounts of unlabelled samples in the target domain. However, in our view, the target of cross-domain few-shot problems is to exploit the \textit{extremely limited samples but not labels} in target domains for recognition. Sufficient data in target domains with only a few labels could also be solved by label propagation techniques, which may not be the real focus in real-world applications. In~\tabref{tab:comp}, other works following the original settings, including NSAE~\cite{liang2021boosting}, TPN-ATA~\cite{wang2021cross} show inferior results than our proposed Dara approach,~\ie, 93.32\% to 95.53\% for 5-shot setting. While NSAE~\cite{liang2021boosting} does not consider the 1-shot setting, which often occurs in some extreme scenarios. In this 1-shot setting, our proposed approach outperforms the newest TPN-ATA~\cite{wang2021cross} by 4.8\% absolute values, which verifies the adaptation abilities of our approach.

\begin{table}[t]
\centering{
\caption{N-way 5-shot classification accuracy on miniImageNet $\to$ BSCD-FSL dataset. Mean accuracies and 95\% confidence interval are reported. The training settings follow the original setting~\cite{guo2020broader} on these datasets.}
\label{table:highwaya}
\setlength{\tabcolsep}{1.6mm}
\renewcommand{\arraystretch}{1.4}
\resizebox{1\linewidth}{!}{
\begin{tabular}{c|c|c|c|c}
\toprule
\textbf{Method}& CropDiseases-10& EuroSAT-All& ISIC-All& ChestX-All\\
\midrule
Finetune~\cite{guo2020broader}&78.04$\pm$0.57&64.25$\pm$0.44&39.56$\pm$0.54&19.18$\pm$0.40\\
BSR~\cite{liu2020feature}&88.62$\pm$0.41&71.71$\pm$0.50&46.94$\pm$0.56&21.44$\pm$0.43\\
TPN-ATA~\cite{wang2021cross}&80.97$\pm$0.49&67.61$\pm$0.36&37.33$\pm$0.43&17.34$\pm$0.28\\
Dara (Ours)&\textbf{91.60$\pm$0.36}&\textbf{77.33$\pm$0.38} &\textbf{49.07$\pm$0.57}& \textbf{21.82$\pm$0.46}\\
\bottomrule
\end{tabular}
}
}
\end{table}

\begin{table}[t]
\centering{
\caption{Ablation studies of 5-way 5-shot learning on three benchmarks. $\mc{M}_{pretrain}$, $\mc{M}_{PFA}$, $\mc{M}_{NDA}$ denotes the source pretraining, prototypical feature alignment and normalized distribution alignment respectively. }
\label{table:ablation}
\setlength{\tabcolsep}{1.2mm}
\renewcommand{\arraystretch}{1.2}
\resizebox{\linewidth}{!}{
\begin{tabular}{ccc|c|c|c}
\toprule
$\mc{M}_{pretrain}$&$\mc{M}_{PFA}$ &$\mc{M}_{NDA}$ & CropDiseases& EuroSAT& ISIC \\
\midrule
Baseline~\cite{guo2020broader}&-&-&87.48$\pm$0.58&75.69$\pm$0.66&43.56$\pm$0.60\\
Ridge&-&-&73.83$\pm$0.51&63.18$\pm$0.77&36.79$\pm$0.59\\
Ridge&-&Mean&92.35$\pm$0.46&81.25$\pm$0.57&49.93$\pm$0.60\\
Ridge&\checkmark&-&93.55$\pm$0.42&80.13$\pm$0.61&51.70$\pm$0.63\\
Ridge&\checkmark&\checkmark&\textbf{95.32$\pm$0.34}&\textbf{85.84$\pm$0.54}&\textbf{56.28$\pm$0.66}\\
\bottomrule
\end{tabular}
}
}
\end{table}

\textbf{Learning with augmented datasets.} Besides the conventional setting, the most straightforward manner for few-shot learning is to extend the small datasets to larger ones, which also shows its effectiveness in some scenarios. To perform fair comparisons with existing methods~\cite{liu2020feature,liang2021boosting,das2021confess,li2022ranking}, we exhibit the results of existing methods and our approach in the bottom half in~\tabref{tab:comp}.
Following the same data augmentation strategies in~\cite{yeh2020large,liu2020feature,liang2021boosting}, our results in the last line show a steady improvement compared to the original setting, which both shows superior results than state-of-the-art methods. In addition, our approach does not use the auxiliary label-propagation techniques as NSAE~\cite{liang2021boosting} and RDC~\cite{li2022ranking}, which utilized the predicted query labels for refining the network parameters.

\textbf{Fine-grained classification for cross-domain few-shot learning.} The fine-grained CDFSL task leads to a different challenge when compared to the vanilla setting~\cite{guo2020broader}. It requires the model not only to have the ability for \textit{generalization} but also the quick adaptation ability to fit the \textit{details}.
To verify the robustness of our proposed approach, here we also adopt the same ResNet-10 backbone following the training protocol in benchmark~\cite{guo2020broader}. As in~\tabref{tab:comp_fine}, it can be found that our model shows clearly leading performance on this fine-grained setting with less domain shifting. For example, in the 5-way 5-shot setting on CUB dataset~\cite{wah2011caltech}, our proposed Dara approach outperforms the fused method BSR~\cite{liu2020feature} by 8.2\%. Despite other works also performing well on some specific datasets, our approach shows a robust performance on these fine-grained datasets.

\textbf{Cross-domain testing with higher ways.} To verify the generalization ability when facing more categories in the target domain, we conduct a performance comparison with three recent methods~\cite{guo2020broader,liu2020feature,wang2021cross} with public codes.~\tabref{table:highwaya} shows the N-way 5-shot settings on the BSCD-FSL dataset. For datasets that contain less than 10 categories, we use all categories for cross-domain testing, denoted as ``-All". Under this setting, our proposed method Dara shows high generalization capability in higher ways compared to existing methods~\cite{guo2020broader,liu2020feature,wang2021cross}.

\subsection{Ablation Study}

\begin{figure}[t]
\centering
\includegraphics[width=\columnwidth]{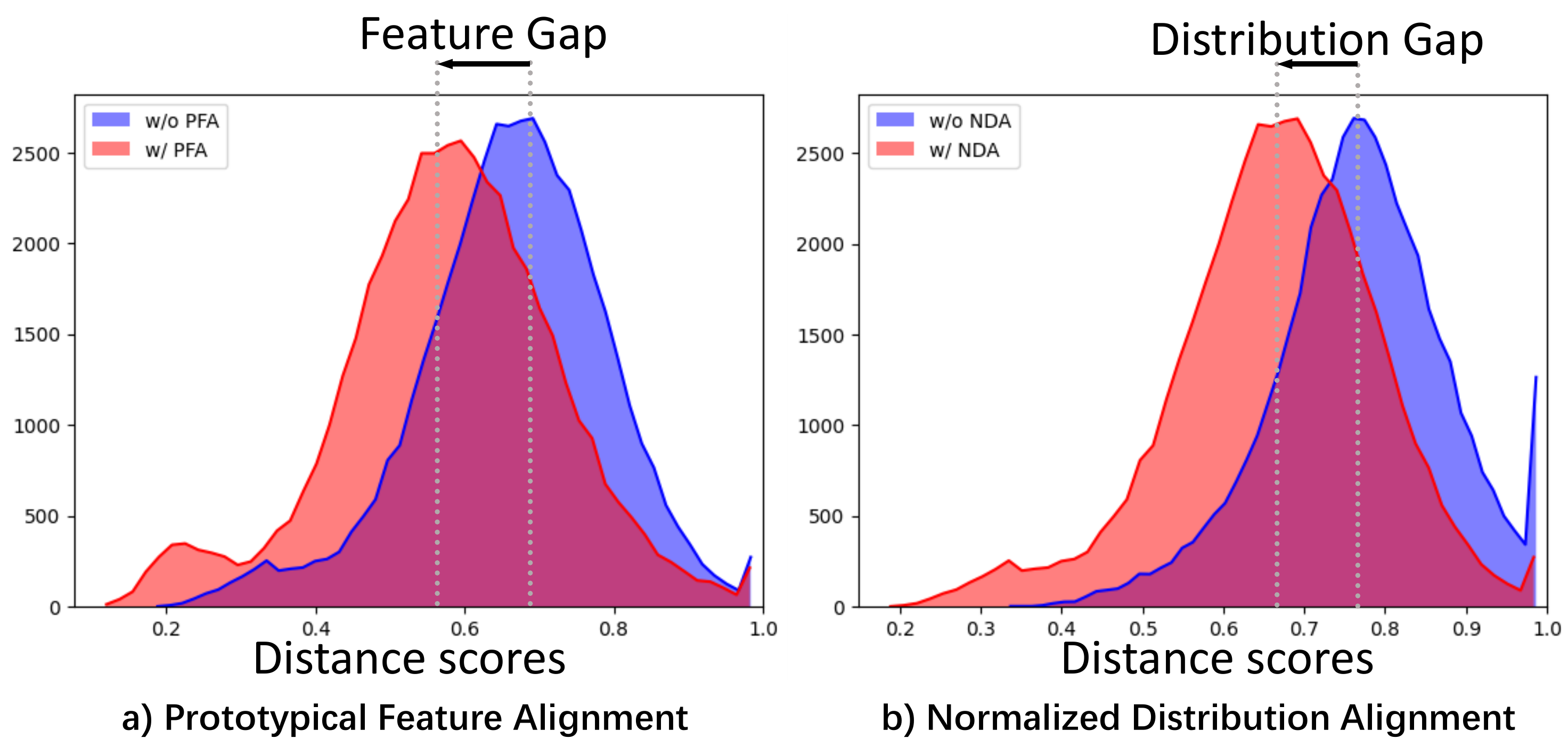}
 \caption{Distance scores of query samples to support samples with the ablation of our proposed two modules. {\color{blue}Blue}: original distance without our module. {\textcolor[RGB]{231,101,26}{Orange}}: with our proposed alignment.
 }
\label{fig:dis_vis}
\end{figure}

\textbf{Effect of dual alignment modules.} To evaluate the effectiveness of our proposed modules, we conduct an ablation study in~\tabref{table:ablation} of our proposed two alignment modules~\ie, prototypical feature alignment $\mc{M}_{PFA}$, and normalized distribution $\mc{M}_{NDA}$. The baseline~\cite{guo2020broader} embedding indicates the vanilla cross-entropy learning, and Ridge denotes learning using constraints in~ Eqn.~\eqref{eq:ridge}. It can be found our two modules works effectively and can fast adapt the existing knowledge to the target queries in the last line. Moreover, we also present the distance scores of query instances to support set in~\figref{fig:dis_vis}, where the left and right subplots visualize the feature and distribution gap with or without our alignments respectively. It can be found that distribution distances with our alignment modules (orange color) are much closer than the original ones.

\begin{figure*}
\begin{center}
\includegraphics[width=1\textwidth]{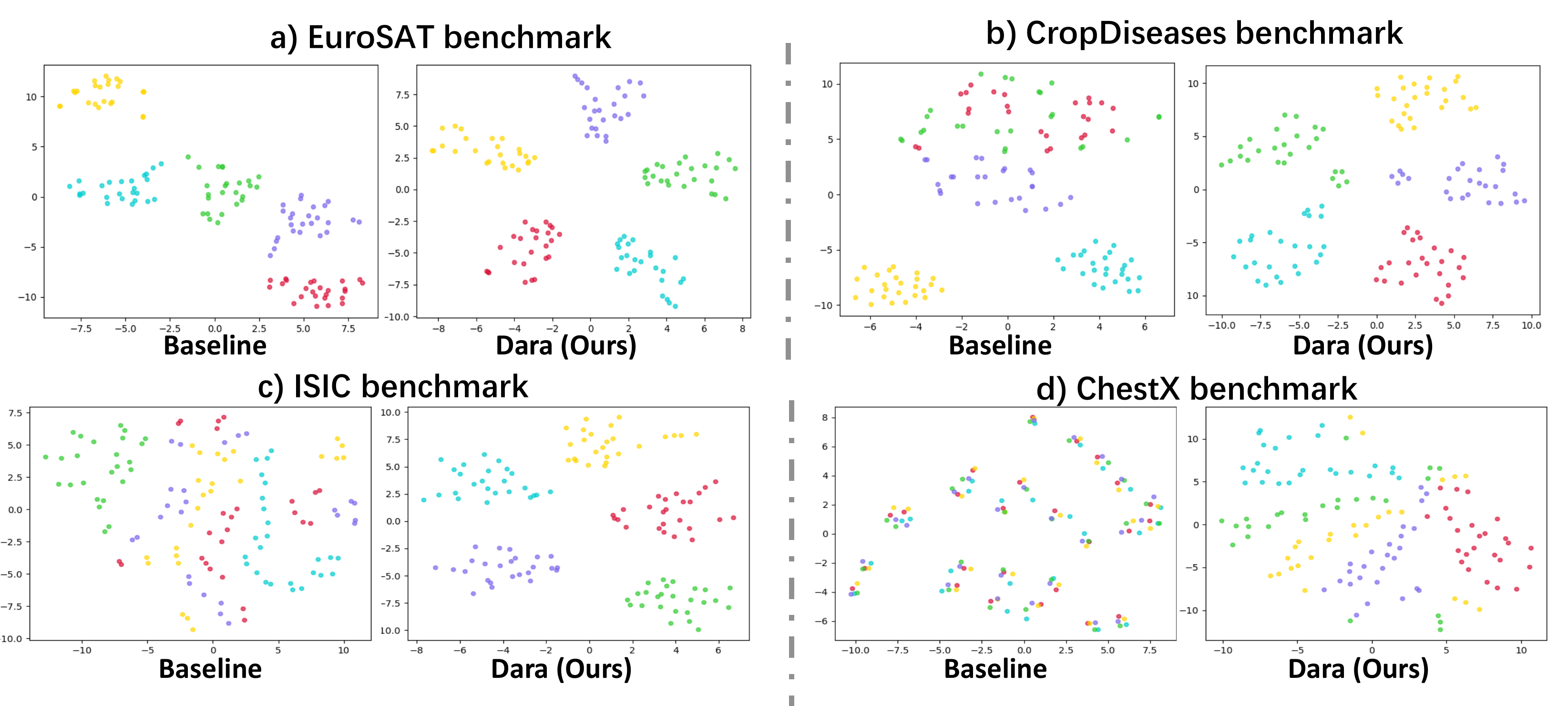}
 \caption{T-SNE visualization of 5-way 5-shot support set embedding. \textbf{Baseline} and \textbf{Ours} are trained with identical training protocols, while \textbf{Baseline} removes prototypical feature and normalized distribution alignments. Each color represents one specific category.  }
 \label{fig:experiment}
 \end{center}
\end{figure*}

\begin{table}[t]
\centering{
\caption{Performances analyses of different statistics and gating strategies after finetuning for the proposed normalized distribution alignment $\mc{M}_{NDA}$. 5-way 5-shot accuracies with confidence intervals are reported. \textbf{S},\textbf{Q} denote the support and query statistics respectively.}
\label{table:norm}
\setlength{\tabcolsep}{0.8mm}
\renewcommand{\arraystretch}{1.4}
\resizebox{1\linewidth}{!}{
\begin{tabular}{c|c|c|c|c}
\toprule
\textbf{Statistics} ($\mu$,$\sigma$) &\textbf{Gating strategy} ($\tau$,1-$\tau$) & CropDiseases& EuroSAT& ISIC \\
\midrule
\textbf{S}&BN (1,0)&93.55$\pm$0.42&80.13$\pm$0.61&51.69$\pm$0.63\\
\textbf{Q}-All&BN (1,0)&94.49$\pm$0.39&81.33$\pm$0.63&51.29$\pm$0.61\\
\textbf{Q}-All&IN (0,1)&93.66$\pm$0.41&81.73$\pm$0.60&52.87$\pm$0.66\\
\textbf{Q}-All&Mean (0.5,0.5)&94.08$\pm$0.40&83.75$\pm$0.55&53.84$\pm$0.64\\
\midrule
\textbf{S}+\textbf{Q}-1&Learnable&94.85$\pm$0.37&84.45$\pm$0.57&55.12$\pm$0.61\\
\textbf{S}+\textbf{Q}-$1\times5$&Learnable& 95.25$\pm$0.33&\textbf{86.06$\pm$0.53}&55.94$\pm$0.65\\
\midrule
\textbf{Q}-All&Learnable&\textbf{95.32$\pm$0.34}&85.84$\pm$0.54&\textbf{56.28$\pm$0.66}\\
\bottomrule
\end{tabular}
}
}
\end{table}

\textbf{Different types of distribution alignment.} Despite the intuition for aligning the statistics with different normalization strategies, we have performed different variants of our proposed distribution alignment modules. As in~\tabref{table:norm}, we fix the gating hyperparameter with all parameters equaling 1 or 0 for simulating the conventional BN and IN operation. After transferring the query statistics from the second line, the performances show a notable improvement compared to the target statistics. Although the average fusion of these two norm features leads to higher performance, the learnable gating strategy shows a steady improvement to final results,~\eg, 2.4\% on the ISIC benchmark dataset.

As our proposed method relies on the querying statistics for distribution alignment, we conduct experiments on three variants of our proposed method at the bottom of~\tabref{table:norm}: 1)\textbf{S}+\textbf{Q}-1: using support statistics with the current one query sample; 2)\textbf{S}+\textbf{Q}-$1\times5$: using support statistics with 1-shot query per class; 3)\textbf{Q}-All: using all query statistics. From \tabref{table:norm}, we find that using 1-shot or only current samples shows comparative performance, which verifies the effectiveness of our proposed distribution alignment module.

\textbf{Effect of prototypical alignment.} In~\tabref{table:proto}, we conduct experiments with four different manners for constructing feature embedding. The first line shows the initial embedding without any alignments. Starting from this initialization, incorporating our prototypical embedding $\br{P}$
 shows an effective improvement for fast domain adaptation. In addition, after performing the instance-level recalibration $\br{I}$, steady improvements can also be made,~\eg, 0.9\% on ISIC based on the high baseline performance.

\textbf{Effect of meta-finetuning strategies.}
To evaluate the effectiveness of our two-stage finetuning strategy, we conduct a detailed ablation in~\tabref{tab:strategy}. Interestingly, with the initialized pretrained measurements in the first line, only optimizing the reprojection weights $\br{Z}$ would lead to over 20\% accuracy improvement on CropDiseases and EuroSAT dataset, which is relatively lightweight for training compared to finetuning the whole backbone network with parameters $\theta$. With the joint optimization in the last line, our model shows a better performance than solely conducting the one-stage optimization. In addition, when taking the backbone features $\theta$ and reprojection weights $\br{Z}$ as a holistic part for joint training, it would lead to bad optimizations due to the reprojection weights $\br{Z}$ being very easy to overfit on the limited training samples. Hence our meta-finetuning strategy follows a stage-wise training manner for optimizing $\theta$ and $\br{Z}$, which is proposed in~\secref{sec:learning}.

\begin{table}[t]
\centering{
\caption{Performances analyses of different feature embedding strategies for the prototypical feature alignment. 5-way 5-shot accuracies with confidence intervals are reported. $\br{P}$: prototypical reprojection. $\br{I}$: instance recalibration.}
\label{table:proto}
\setlength{\tabcolsep}{1.0mm}
\renewcommand{\arraystretch}{1.4}
\resizebox{1\linewidth}{!}{
\begin{tabular}{c|c|c|c|c}
\toprule
$\mc{M}_{NDA}$ & \textbf{Feature Embedding} & CropDiseases& EuroSAT& ISIC \\
\midrule
-&Init.&73.83$\pm$0.51&63.18$\pm$0.77&36.79$\pm$0.59\\
-&$\br{P}$&93.55$\pm$0.42&80.13$\pm$0.61&51.69$\pm$0.63\\
\checkmark&$\br{P}$&95.25$\pm$0.35&85.46$\pm$0.52&55.40$\pm$0.65\\
\checkmark& $\br{I}+\br{P}$&\textbf{95.32$\pm$0.34}&\textbf{85.84$\pm$0.54}&\textbf{56.28$\pm$0.66}\\
\bottomrule
\end{tabular}
}
}
\end{table}

\begin{table}[t]
\centering{
\caption{Performances analyses of different meta finetuning strategies for the prototypical feature alignment. 5-way 5-shot accuracies with confidence intervals are reported. $\theta$ Opt.: optimizing the meta-learner backbones $\theta$. $\br{Z}$ Opt.: only optimizing the prototypical reprojection weight $\br{Z}$.}
\label{tab:strategy}
\setlength{\tabcolsep}{2.5mm}
\renewcommand{\arraystretch}{1.4}
\resizebox{1\linewidth}{!}{
\begin{tabular}{c|c|c|c|c}
\toprule
\multicolumn{2}{c|}{\textbf{Meta Strategy}} & \multicolumn{3}{c}{\textbf{Datasets}} \\
\cline{1-5}
$\theta$ Opt.& $\br{Z}$ Opt.& CropDiseases& EuroSAT& ISIC \\
\midrule
-&-&73.83$\pm$0.51&63.18$\pm$0.77&36.79$\pm$0.59\\
-&\checkmark&94.42$\pm$0.38&83.94$\pm$0.56&53.26$\pm$0.60\\
\checkmark&-&94.87$\pm$0.40&85.20$\pm$0.51&55.90$\pm$0.63\\
\checkmark&\checkmark&\textbf{95.32$\pm$0.34}&\textbf{85.84$\pm$0.54}&\textbf{56.28$\pm$0.66}\\
\bottomrule
\end{tabular}
}
}
\end{table}

\begin{table*}[t]
\setlength{\tabcolsep}{4.0mm}
\caption{5-way k-shot classification accuracy on miniImageNet $\to$ BSCD-FSL dataset. Mean accuracies and 95\% confidence interval are reported. The training settings follow the original setting~\cite{guo2020broader} on these datasets. Query vs. Support: the partition of pseudo query and pseudo support during meta-finetuning. Higher performance in the same setting is bolded for a better view.}
\label{tab:rate}
\renewcommand{\arraystretch}{1.3}
\centering
\resizebox{\textwidth}{!}{
\begin{tabular}{c cc|cc|cc|cc}
\toprule[1.5pt]
\multirow{1}{*}{Query vs. Support}& \multicolumn{2}{c}{CropDiseases}& \multicolumn{2}{c}{EuroSAT} & \multicolumn{2}{c}{ISIC} & \multicolumn{2}{c}{ChestX} \\
\cline{2-9}
$\{ \widetilde{\mc{T}}^{pq},\widetilde{\mc{T}}^{ps} \}$& 5-shot& 20-shot& 5-shot& 20-shot& 5-shot& 20-shot & 5-shot& 20-shot \\
\midrule[1.5pt]
$\{5\%, 95\% \}$ & N/A&\textbf{98.58$\pm$0.15} &N/A& \textbf{93.14$\pm$0.30}&N/A& 67.36$\pm$0.57 &N/A&\textbf{35.95$\pm$0.47} \\
$\{20\%, 80\% \}$ &\textbf{95.32$\pm$0.34}&98.57$\pm$0.15 &\textbf{85.84$\pm$0.54}& 92.11$\pm$0.33&\textbf{56.28$\pm$0.66}& \textbf{69.12$\pm$0.57}&\textbf{27.54$\pm$0.42} &34.79$\pm$0.46 \\
$\{40\%, 60\% \}$ &94.90$\pm$0.37&98.27$\pm$0.17 &85.46$\pm$0.56& 91.85$\pm$0.33&56.03$\pm$0.67& 68.10$\pm$0.59&27.03$\pm$0.43 &34.20$\pm$0.50 \\
$\{60\%, 40\% \}$ &95.02$\pm$0.36&98.18$\pm$0.19 &85.14$\pm$0.55& 92.04$\pm$0.33&56.05$\pm$0.64& 68.12$\pm$0.58&26.90$\pm$0.43 &33.60$\pm$0.47 \\
$\{80\%, 20\% \}$ &94.72$\pm$0.36&98.13$\pm$0.19 &84.15$\pm$0.56& 91.50$\pm$0.33&54.71$\pm$0.61& 68.03$\pm$0.59&26.43$\pm$0.44 &33.16$\pm$0.47 \\
\bottomrule[1.5pt]
\end{tabular}
}
\end{table*}

\subsection{Performance Analysis}
\textbf{Pseudo sampling strategies for meta-finetuning.}
One obvious issue for meta-finetuning is that the pseudo query and pseudo support samples from $\{ \widetilde{\mc{T}}^{pq},\widetilde{\mc{T}}^{ps} \}$ are partitioned by the target query samples. Hence the partition rate of query and support samples are thus important in the meta-learning phase.~\tabref{tab:rate} exhibits the different rates of query samples vs. support samples. Note that the 5-shot setting only consists of 5 samples for learning, thus the 5\% partition is not applicable. It can be found that with fewer samples as the support set, the performance shows an improvement, especially on the ``hard" dataset,~\eg, ChestX dataset with medical gray images. Moreover, as discovered both theatrically and empirically, the query and support set must be disjointed, otherwise, the Ridge Regression of $\br{W}^*= \mathop{\arg\min}_{\br{W}} ||\br{W}\br{P}_{i}- \br{Q} ||^2 + \lambda ||\br{W}||^2$ would be meaningless when $\br{P} \equiv \br{Q}$.

\textbf{Resolutions for cross-domain feature reconstruction.} In our prototypical feature reconstruction module, we maintain the spatial resolutions for representation learning instead of the conventional pooling operation. This leads to an interesting exploration in~\tabref{table:resolution}. It is observed that \textit{easy} dataset (\ie, CropDiseases) requires dense features while the \textit{difficult} dataset (\ie, ChestX) shows better performance with sparse features. This indicates that our proposed method is adaptive when tackling datasets with different difficulties.

\textbf{Normalization rate for cross-domain feature reconstruction.} The other important factor in prototypical feature reconstruction is the normalization rate,~\ie, $\lambda=\frac{KR}{C} \times \beta$, here $\beta$ denotes the normalization scales related to the reconstruction process. As in~\tabref{table:beta}, the normalization process is indispensable but not sensitive during the training phase. We empirically set $\beta=1$ for all experiments in this work.

\begin{table}[t]
\centering{
\caption{Performances analyses of different resolutions in feature space. 5-way 5-shot accuracies with confidence intervals are reported.}
\label{table:resolution}
\setlength{\tabcolsep}{1.4mm}
\renewcommand{\arraystretch}{1.4}
\resizebox{1\linewidth}{!}{
\begin{tabular}{c|c|c|c|c}
\toprule
\textbf{Feature Resolution $R$}& CropDiseases& EuroSAT& ISIC& ChestX \\
\midrule
$(3\times 3)$&93.27$\pm$0.42&84.80$\pm$0.50&53.95$\pm$0.67&\textbf{27.57$\pm$0.44}\\
$(4\times 4)$&94.33$\pm$0.41&85.45$\pm$0.53&55.50$\pm$0.61&27.45$\pm$0.44\\
$(5\times 5)$&95.32$\pm$0.34&\textbf{85.84$\pm$0.54}&\textbf{56.28$\pm$0.66}&27.54$\pm$0.42\\
$(6\times 6)$&95.35$\pm$0.35&85.05$\pm$0.56&55.97$\pm$0.66&26.34$\pm$0.43\\
$(7\times 7)$&\textbf{95.38$\pm$0.35}&84.48$\pm$0.54&56.07$\pm$0.69&26.00$\pm$0.43\\
\bottomrule
\end{tabular}
}
}
\end{table}

\begin{table}[t]
\centering{
\caption{Performances analyses of different normalization rates $\lambda=\frac{KR}{C} \times \beta$. 5-way 5-shot accuracies with confidence intervals are reported.}
\label{table:beta}
\setlength{\tabcolsep}{1.4mm}
\renewcommand{\arraystretch}{1.4}
\resizebox{1\linewidth}{!}{
\begin{tabular}{c|c|c|c}
\toprule
\textbf{Normalization Rate }& CropDiseases& EuroSAT& ISIC \\
\midrule
$\beta=0$&95.03$\pm$0.38&82.38$\pm$0.50&55.45$\pm$0.63\\
$\beta=0.1$&95.01$\pm$0.36&84.49$\pm$0.52&56.09$\pm$0.64\\
$\beta=1$&\textbf{95.32$\pm$0.34}&\textbf{85.84$\pm$0.54}&\textbf{56.28$\pm$0.66}\\
$\beta=10$&95.27$\pm$0.35&85.54$\pm$0.52&55.99$\pm$0.68\\
\bottomrule
\end{tabular}
}
}
\end{table}

\begin{figure*}[t]
\begin{center}
\includegraphics[width=1\textwidth]{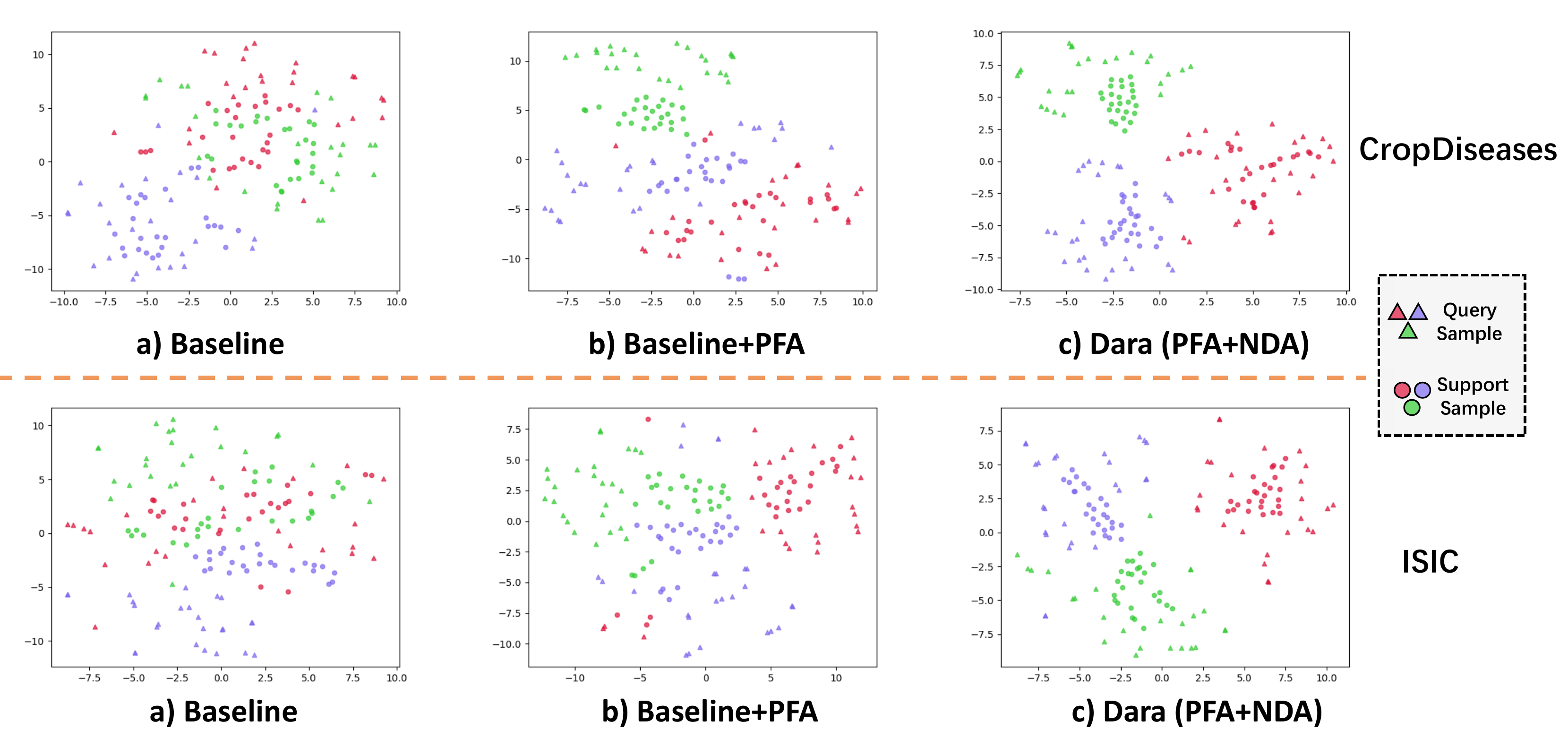}
 \caption{T-SNE visualization of K-way support and query joint embedding. a) \textbf{Baseline} without prototypical feature alignment (PFA) and normalized distribution alignment (NDA). b) \textbf{Baseline} model with PFA module. c) \textbf{Ours} with PFA and NDA modules. $K=3$ are selected for better view.   }
 \label{fig:app2}
 \end{center}
\end{figure*}

\textbf{Visualization for fast cross-domain adaptations.} Besides the quantitative verifications, we also conduct a visualization of support samples in~\figref{fig:experiment} for a better understanding of our alignment approach. We first exhibit the support samples on two \textit{easy} datasets which share similarities with the natural images,~\ie, CropDiseases and EuroSAT in~\cite{guo2020broader}. It can be found that both baseline and Dara approach shows distinguishable embedding on this 2-dimension space using t-SNE visualization, but the baseline model faces difficulties on the CropDiseases dataset.
In the second row, we exhibit two \textit{difficult} datasets,~\ie, ISIC and ChestX, which show the most dissimilar visual appearances with natural source data. The baseline model denotes the identical training protocol to our model but removes the two alignment modules. These visualized cluster exemplars show that our approach (Dara) has a strong potential for distinguishing hard instances in extreme domain-shifting scenarios.

\textbf{Joint distribution visualization on category clusters.} To verify the joint distribution of both query and support samples, we present a 5-way X-shot setting in~\figref{fig:app2} with randomly selected three categories for a better view. The triangles and circles represent the query samples and support samples respectively. Besides the visualizations on support samples above, this joint distribution provides evidence for aligning the query and support samples in the feature space. It can be found with the proposed prototypical feature alignment (PFA) and normalized distribution alignment (NDA) modules, the query samples, and their referred support samples are transformed into the same clusters, which shows great benefits for further few-shot measurements.

\subsection{Limitations and Future Work}
Although promising results have been achieved in this proposed approach, it should be noted that the challenges are still huge in the cross-domain setting. For example, the best model performs less than 30\% accuracy for the 5-shot setting on the most difficult ChestX dataset~\cite{wang2017chestx}. This reveals that common knowledge like ImageNet~\cite{deng2009imagenet} can only provide a reasonable distribution for initialization, but it is very hard to learn the real \textit{expert knowledge} in some medical applications. Moreover, the setting we explored is still under the $N$-way $K$-shot learning system, while the real-world demands often require an adaptive $X$-way or $Y$-shot for both learning and inference, which should also be explored in future work. We believe that this could be solved by learning adaptive reprojections and alignment strategies which are highly related to input instances.

\section{Conclusion}\label{sect:conclusion}
In this paper, we propose to explore the cross-domain few-shot learning problems with only a few samples available in the target domain. Different from the existing methods focusing on domain generalizations, we propose to fast adapt the existing knowledge to the target queries with a dual adaptive representation alignment approach, which consists of two essential modules. The first module focuses on reprojecting the support instances and prototypes to the feature space of target queries by a differential closed-form solution, which can perform a fast feature-wise transformation with few additional parameters.
The second module introduces a new normalized distribution alignment strategy that aligns the existing base knowledge to the distribution of target queries by calculating the statistical priors. With these two efficient modules, we further present a progressive meta-learning framework, which constrains the generalization of feature learning and performs a fast adaptation to the final recognition space simultaneously.

\ifCLASSOPTIONcompsoc

  \section*{Acknowledgments}
\else

  \section*{Acknowledgment}
\fi
This work is partially supported by grants from the National Natural Science Foundation of China under contracts No. 62132002, No. 61825101, and No. 62202010 and also supported by China Postdoctoral Science Foundation No. 2022M710212.

\ifCLASSOPTIONcaptionsoff
  \newpage
\fi

\balance
\bibliographystyle{IEEEtran}
\bibliography{fewshot}

\end{document}